%% file: main.tex
\documentclass{article}
\usepackage[utf8]{inputenc}
\usepackage[margin=1.25in]{geometry}

\usepackage{amsfonts}
\usepackage{amsthm}
\usepackage{amsmath}
\usepackage{amssymb}

\PassOptionsToPackage{sort}{natbib}
\PassOptionsToPackage{square}{natbib}
\usepackage{natbib}

\newif\ifcvpr

\usepackage{xfrac} 
\usepackage{nicefrac}
\usepackage{xspace}
\usepackage[skip=0pt]{subcaption}
\usepackage{url}
\usepackage{hyperref}
\hypersetup{
    colorlinks=true,
    linkcolor=blue,
    filecolor=blue, 
    urlcolor=magenta,
    citecolor=blue,
    }

\usepackage[ruled,vlined]{algorithm2e}
\usepackage{multicol}
\usepackage{multirow}
\usepackage{siunitx}
\usepackage{tablefootnote}

\usepackage[colorinlistoftodos,prependcaption]{todonotes}

\definecolor{darkgreen}{rgb}{0,0.4,0.0}

\makeatletter
\@ifundefined{theorem}{%
}{}
\@ifundefined{lemma}{%
}{}
\@ifundefined{corollary}{%
}{}
\@ifundefined{conjecture}{%
}{}

\PassOptionsToPackage{usenames,dvipsnames}{xcolor}
\usepackage{cleveref}

\newcommand{\fedavg}{{FedAvg}\xspace}
\newcommand{\fedsgd}{{FedSGD}\xspace}

\newcommand{\ouralg}{DP-FedEmb\xspace}
\newcommand{\ourmetric}{recall@FAR=$1\mathrm{e}{-3}$\xspace}
\newcommand{\Ourmetric}{Recall@FAR=$1\mathrm{e}{-3}$\xspace}
\newcommand{\data}{\ensuremath{\mathcal{D}}}
\newcommand{\userset}{\ensuremath{\mathcal{U}}}
\newcommand{\vcset}{\ensuremath{\mathcal{S}}}
\newcommand{\serveropt}{\textsc{ServerOpt}\xspace}
\newcommand{\clientopt}{\textsc{ClientOpt}\xspace}

\renewcommand\cite{\citep}
\newcommand\eg{e.g.}
\newcommand\ie{i.e.}

\usepackage[affil-sl]{authblk}

\makeatletter
\renewcommand\AB@affilsepx{,\quad \protect\Affilfont}
\makeatother

\title{Learning to Generate Image Embeddings \\ with User-level Differential Privacy}
\author{Zheng Xu\thanks{The first two authors contributed equally. Correspondence to Zheng Xu \texttt{xuzheng@google.com}}}
\author{Maxwell Collins$^*$} 
\author{Yuxiao Wang} 
\author{Liviu Panait} 
\author{Sewoong Oh}
\author{Sean Augenstein}
\author{Ting Liu}
\author{Florian Schroff}
\author{H. Brendan McMahan}
\affil{Google Research}
\date{}

\begin{document}
\setboolean{cvpr}{false} 

\maketitle

\input{sec_abstract}

\input{sec_intro}

\input{sec_related}

\input{sec_method}

\input{sec_exp}

\input{sec_conclusion}

\input{sec_ack}

\bibliographystyle{plainnat}
\bibliography{refer}

\end{document}

%% file: sec_abstract.tex
\begin{abstract}
Small on-device models have been successfully trained with user-level differential privacy (DP) for next word prediction and image classification tasks in the past. However, existing methods can fail when directly applied to learn embedding models using supervised training data with a large class space.
To achieve user-level DP for large image-to-embedding feature extractors, we propose \ouralg, a variant of federated learning algorithms with per-user sensitivity control and noise addition, to train from user-partitioned data centralized in the datacenter. \ouralg combines virtual clients,  partial aggregation, private local fine-tuning, and public pretraining to achieve strong privacy utility trade-offs. We apply \ouralg to train image embedding models for faces, landmarks and natural species, and demonstrate its superior utility under same privacy budget on benchmark datasets DigiFace, EMNIST, GLD and iNaturalist. We further illustrate it is possible to achieve strong user-level DP guarantees of $\epsilon<4$ while controlling the utility drop within 5\%, when millions of users can participate in training .
\end{abstract}

%% file: sec_intro.tex
\section{Introduction}

Representation learning, by training deep neural networks as feature extractors to generate compact embedding vectors from images, is a fundamental component in computer vision. Metric learning, a kind of representation learning using supervised data, has been widely applied to image recognition, clustering, and retrieval \citep{weinberger2009distance,schroff2015facenet,weyand2020google}.
Machine learning models have the capacity to memorize training data  \citep{carlini2019secret,carlini2021extracting}, leading to privacy risks when the models are deployed. Privacy risk can also be audited by membership inference attacks \citep{shokri2017membership,carlini2022membership}, i.e. detecting whether certain data was used to train a model and potentially exposing users' usage behaviors. Defending against such risks is a critical responsibility when training on privacy-sensitive data. 

Differential Privacy (DP) \citep{dwork2006calibrating} is an extensively used quantifiable measurement of privacy risk, now generally accepted as a standard notion of privacy in both industry and government \citep{apple,ding2017collecting,bureau2021disclosure,dpftrl_blogpost}. 
Applied to machine learning, DP requires a training procedure with explicit randomness, and guarantees that the distribution over output models is quantifiably similar given a certain scope of change to the training dataset.
A DP guarantee with respect to the change of a single arbitrary training example is known as  \emph{example-level DP}, which provides plausible deniability (in the binary hypothesis testing sense of \cite{kairouz2015composition}) that any single example (e.g., image) occurred in the training dataset. If we instead consider how the distribution of output models changes if the data (including even the number of examples) from any single user change arbitrarily, we have \emph{user-level DP} \citep{dwork2010differential}. This ensures model training is quantifiably insensitive to all of the data from any one user, and hence it is impossible to tell if a user has participated in training with high confidence. This guarantee can be exponentially stronger than example-level DP if one user may contribute many examples to training.

Recently, DP-SGD \citep{abadi2016deep} (essentially, SGD with the additional steps of clipping each individual gradient to have a maximum norm, and adding correspondingly calibrated noise) has been used to achieve example-level DP for relatively large models in language modeling and image classification tasks \citep{li2021large,yu2021differentially,anil2021large,de2022unlocking,kurakin2022toward},  often utilizing techniques like large batch training and pretraining on public data.
DP-SGD can be modified to guarantee user-level DP, which is often combined with federated learning algorithms and called DP-\fedavg \citep{mcmahan18learning}. User-level DP has only been studied for small on-device models that have less than 10 million parameters \citep{mcmahan18learning,kairouz21practical,ramaswamy2020training}. 

\begin{figure*}[htb]
\centering
\includegraphics[width=0.85\textwidth]{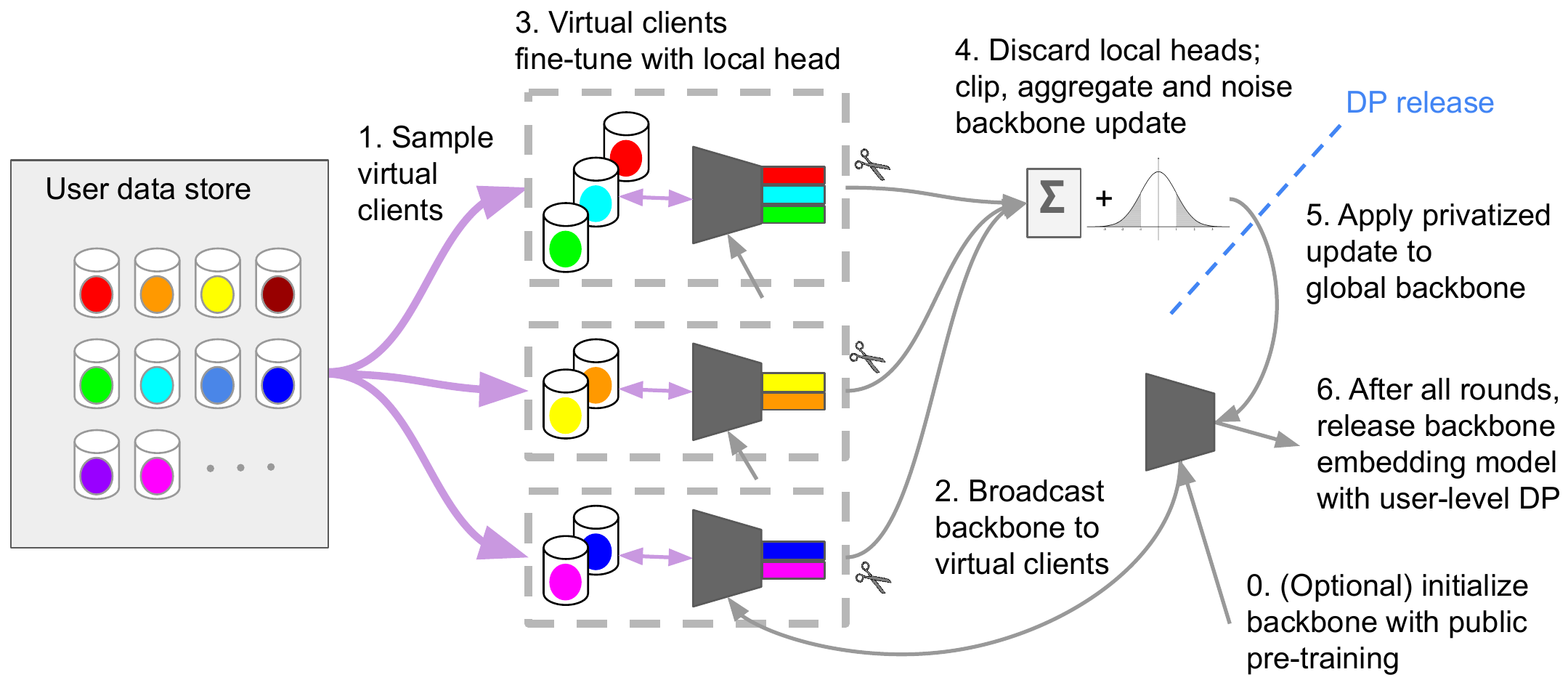}
\caption{\ouralg combines virtual clients, local fine-tuning, partial aggregation, and public pretraining to achieve strong privacy utility trade-offs. Colors indicate different users and the simplified case of a single class per user, coloring the softmax head accordingly.} 
\label{fig:dp-fedemb-teaser}
\ifthenelse{\boolean{cvpr}}{\vspace{-0.3cm}}{}
\end{figure*}

We consider user-level DP for relatively large models in representation learning with supervised data. In our setting, similar to federated learning (FL), the data are user-partitioned; but in contrast to decentralized FL, we are primarily motivated by centralized data that benefit from access to richer computation resources and the ability to form virtual clients at random. 
Throughout this work, we use \emph{user} as the basic unit of data partitioning and the granularity for privacy; a user owns their (image) data, and \emph{class}, \emph{identity} and \emph{label} are used interchangeably for the supervised information. 

Though typically each user only contributes images for a small number of classes, the combined class space from the union of all users can be very large, which proves challenging for existing DP algorithms. When the model size is fixed independent of the number of users, and at a relatively small scale with a few million parameters, previous methods can only achieve strong user-level DP when millions of users are available \citep{mcmahan18learning,kairouz21practical,ramaswamy2020training}.
In contrast, when considering learning embedding models for applications like facial images, the number of classes (and hence, the total model size) can grow linearly with the number of users, and so simply scaling up to larger datasets with more users no longer ensures that good privacy-utility trade-offs can be achieved. For example, in a standard multi-class training paradigm with $128$-dimensional embedding vectors, with one million users we expect the final dense layer for prediction alone to have over $128$ million trainable parameters.  Further, the fact that most users will only have examples from a small number of classes implies that gradients are approximately sparse, whereas DP-SGD requires the addition of dense noise to the full gradient, leading to a poor signal-to-noise ratio in the updates. Hence, existing methods can easily fail on the problems we consider.

We propose \ouralg to train embedding models with user-level differential privacy; \cref{fig:dp-fedemb-teaser} provides a high-level overview.  \ouralg combines  public pretraining, virtual clients, local fine-tuning, and partial aggregation to achieve strong privacy-utility trade-offs. The key to the approach is partitioning the model into a backbone network that generates embeddings, and a classification softmax head specific to the classes in the training data. In each training round, users are grouped into virtual clients and initialized from the global backbone. A local randomly-initialized softmax head layer is added for the limited number of classes on the virtual client, and the complete local model is fine-tuned in order to produce an update to the backbone. The local head parameters are not included in the private aggregation and hence require no noise addition. This is in contrast to existing methods like DP-FedAvg/DP-SGD, which would add noise to all parameters including the softmax head. The backbone updates are clipped to a maximum L2 norm, aggregated across virtual clients, and combined with appropriate DP noise. At this point, the noised update is the output of a DP mechanism and satisfies the corresponding DP guarantee. This update is then applied to the global backbone, which inherits the DP guarantee, and passed to the next round of training. 
\ouralg significantly improves the scalability of DP training for embedding models, as only the parameters of the backbone network are privatized and released, and the size of this portion of the model does not grow with the number of users. Pretraining the backbone network on public data to learn general visual representations before applying \ouralg for more privacy sensitive tasks further improves performance. 

We demonstrate the superior performance of \ouralg for embedding models by experiments on datasets with moderate size of users and classes (DigiFace of $98.96K$ or $9047$ users$\,/\,$identities, Google Landmarks Dataset of 1262 users and 2028 classes, and iNaturalist of 9275 users and 1203 classes). We also show relatively strong privacy guarantees of single digit $\epsilon$ can be achieved while maintaining strong utility if millions users can participate in training. To our knowledge, this is the first report of training a commonly used large vision model, ResNet-50, with non-negligible noise for user-level DP.






%% file: sec_related.tex
\section{Related work}

\paragraph {Differential Privacy (DP),} introduced by \cite{dwork2006calibrating}, is a formal mathematical notion of privacy protection. Formally, two datasets $D$ and $D'$ are said to be neighboring if they differ at most by one entry. A randomized mechanism ${\cal A}$ is said to be $(\epsilon,\delta)$-differentially private if $ {\mathbb P}({\cal A}(D)\in S) \leq e^\epsilon {\mathbb P}({\cal A}(D')\in S) +\delta$, for all neighboring $D$ and $D'$. We refer to this original definition as {\em example-level DP}, and several variations have been proposed, including Renyi DP (RDP) \citep{mironov2017renyi}, Privacy Loss Distribution (PLD) \citep{koskela2020tight,doroshenko2022connect}, and concentrated DP (zCDP) \citep{bun2016concentrated}. 

 A formal definition of {\em user-level DP} is introduced in  \cite{dwork2010differential}, where the unit of privacy protection is extended from a single entry (in the original example-level DP) to every entry that belongs to the same user. The dependence of the utility on the number of users and the number of samples per user has been studied for various tasks: empirical risk minimization and mean estimation \citep{levy2021learning}, 
 estimating discrete distributions  \citep{liu2020learning}, and PAC learning  \citep{ghazi2021user}. Extensions to heterogeneous users in sample size have been studied in \citep{amin2019bounding,epasto2020smoothly}. 
User-level DP is particularly useful in federated learning, where the natural unit of privacy is a user (i.e., a client) \citep{geyer2017differentially,mcmahan2018general} and standard private training algorithms respect user-level DP \citep{mcmahan18learning}. 
The privacy-utility trade-off for user-level DP is investigated in \citep{jain2021differentially,shenshare} where feature extractors are federated trained and classifiers are personalized. 

\paragraph{Representation learning} and metric learning are active research directions in computer vision. Recently, a lot of progress has been made towards representation learning with large-scale unsupervised data \citep{chen2020simple,he2022masked,zbontar2021barlow,grill2020bootstrap}. However, we consider representation learning with supervised data as it is widely used for downstream tasks like face recognition and clustering \citep{schroff2015facenet,taigman2014deepface}, person re-identification \citep{wu2016personnet}, and landmark recognition \citep{weyand2020google}, which can significantly benefit from privacy protection. Two technical frameworks are often used in the supervised representation learning tasks due to the large output space: triplet and its variants with hard negative mining \citep{schroff2015facenet}, and multi-class training with proxy weights \citep{taigman2014deepface,deng2019arcface,wen2021sphereface2}. We propose \ouralg based on the multi-class approach for the following reasons: the two approaches can achieve similar performance when trained with large-scale data \citep{musgrave2020metric}; multi-class training is simple and flexible, and can be more efficient when less data are touched every iteration; and negative sampling can trigger non-trivial computational and privacy cost. To the best of our knowledge, differentially private models have not been trained for large-scale representation learning.

\paragraph{Federated learning} is an active research topic primarily designed for learning from decentralized data \citep{kairouz2019advances,wang2021fieldguide}. We 
propose \ouralg based on federated learning algorithms as they are suitable for user-level DP. User-level DP can be achieved in federated learning by variants of DP-\fedavg \citep{geyer2017differentially,mcmahan18learning,kairouz21practical,ramaswamy2020training,andrew2019differentially};  the previous works train relatively small models for image classification and language modeling tasks. Closest to \ouralg is 
\citet{singhal2021federated}, which proposed federated reconstruction that performs partially local training for personalization; \citet{dong2022spherefed} fixed the softmax and train the feature extractor before calibrating for image classification tasks; \citet{waghmare2022efficient} modified sampled softmax for large output space in federated learning. These works are designed for learning with decentralized data, and do not consider differential privacy. \citet{meng2022improving} use differential privacy on proxy vectors to mitigate the privacy concerns when clients exchange weight vectors of identities for federated training, which is different from our motivation of training a differentially private model that will not memorize a specific user's data.

%% file: sec_method.tex
\section{Learning Embedding Models}

\subsection{Problem formulation and centralized training} 
We learn a \emph{backbone} network  parameterized by $\theta$, that outputs an embedding vector $z=f(\theta, x)$ for an input image $x$. The backbone network, $\theta$, is trained on paired examples of image $x$ and class $y$. The training dataset is naturally partitioned by $M$ users, i.e., $\data = \bigcup_{i=1}^M \data_i$. 

We adopt the popular multi-class training framework for embedding models, where a proxy weight vector $w_j$ is learnt for each class $j$. We use $\omega$ to denote the union of proxy weight vectors $\{w_j\}$, which is called the \emph{head} of the network. Given a training image-class pair $(x, y)$, logits are computed by taking the inner product between the embedding vector $f(\theta, x)$ and the proxy weight vectors in the network head $\omega$. This is effectively passing $f(\theta, x)$ through a dense network layer parameterized by $\omega$ without the bias terms. With a supervised training loss $\ell$, such as the cross entropy loss, the following objective is optimized
\begin{align} \label{eq:cen_obj}
    \min_{\theta, \omega} \;\;  \sum_{(x,y)\in \data} \, \ell(\langle \omega, \, f(\theta, x)\rangle, \, y) \,,
\end{align}
where we overload the inner product notation, i.e., $\langle \omega, \, f(\theta, x)\rangle$, to denote a set of inner products with each element in $\omega$. 
Typically, 
variants of the gradient descent method are used to solve the optimization in  \eqref{eq:cen_obj}.
In each iteration, the average gradient is computed on a sampled minibatch of data $B \subset \data$, and is then used to update the model parameters $\theta$ and $\omega$.
Furthermore, when the output space, i.e., the number of classes, is very large, sampled softmax is often applied \citep{jean2014using}, where only a subset of proxy weights sampled from $\omega$ are used in each training iteration.

\subsection{User-level DP and DP-\fedavg} \label{sec:dp-fedavg} 

\ifthenelse{\boolean{cvpr}}{}{\input{dp_fedavg_alg}}

To achieve user-level DP, we control the sensitivity of each user and add corresponding noise for anonymization. To effectively control the sensitivity, it is important to understand and account for the contributions of each user in the model updates; hence it is convenient to consider the data at a granularity of users instead of individual samples. Grouping together each user's data, the objective \eqref{eq:cen_obj} can be rewritten as 
\begin{align} \label{eq:fed_obj}
\min_{\theta, \omega}\;\;  \sum^{M}_{i=1} \sum_{(x,y)\in \data_i} \, \ell(\langle \omega, \, f(\theta, x)\rangle, \, y) \,.
\end{align}
The above objective  of two level sum is often found in federated learning  \citep{wang2021fieldguide}, which can be optimized by the (generalized) FedAvg algorithm \citep{mcmahan2017fedavg,reddi2021adaptive}. In generalized FedAvg, each round $t$ starts with the server broadcasting $\theta^{(t)}, \omega^{(t)}$ to a subset of clients. Each client $i$ will then update the local model parameters by  \clientopt with private data $\data_i$, and send back the updates for model parameters $\Delta_i(\theta^{(t)}), \Delta_i(\omega^{(t)})$. The model deltas from sampled clients are then aggregated and used by \serveropt to get $\theta^{(t+1)}, \omega^{(t+1)}$ for the next round.

The generalized \fedavg algorithm can be extended for user-level DP by clipping the model deltas and adding noise proportional to the sensitivity \citep{mcmahan18learning,geyer2017differentially}. We can use either independent Gaussian noise \citep{mcmahan18learning}, or correlated noise that can achieve comparable privacy-utility trade-off without relying on the assumption of sampling \citep{kairouz21practical}. The two variants are effectively applying DP-SGD \citep{abadi2016deep} or DP-FTRL \citep{kairouz21practical} as \serveropt in the generalized \fedavg framework. Unlike the cross-device FL setting where sampling is extremely hard, it is possible to control user sampling in the datacenter and use DP-SGD. But DP-FTRL provides the possibility of handling the online setting where the user data are streamed instead of collected, and can be accounted for zCDP \citep{bun2016concentrated} reported by US census bureau \citep{bureau2021disclosure}. A complete description of DP-\fedavg for training a backbone network to generate image embeddings is in \cref{algo:dp-fedavg} \ifthenelse{\boolean{cvpr}}{in \cref{app:fedavg}{}} .

In addition to the flexibility of generalized \fedavg for user-level DP, there are a few side effects of \fedavg that make it particularly effective for differentially private training. The model deltas are computed based on data for each user before clipping in DP, which can potentially reduce the bias introduced by clipping. As \citet{de2022unlocking} suggested that averaging gradients from augmented data before clipping can improve training for example-level DP, model deltas from user data for user-level DP can be considered a natural extension to improve the bias-variance trade-off. The communication efficiency of \fedavg that leads to infrequent aggregation and model release is also desirable for DP training. The local model updates by private data on clients introduce no additional privacy cost, and only communication rounds between clients and server have to be accounted for DP. Though the theoretical advantages of \fedavg are only proved under certain assumptions \citep{woodworth2020local,wang2022unreasonable}, \fedavg with local updates can achieve communication efficiency and fast convergence in various practical applications \citep{wang2021fieldguide}.

\subsection{Proposed \ouralg method} \label{sec:dp-fedemb}

While generalized DP-\fedavg can be applied to train a backbone network $\theta$ to generate embedding $f(\theta, x)$ from image $x$, there are challenges that significantly affect the efficiency and feasibility of the method. We propose \ouralg with a few key features: virtual clients, partial aggregation, local fine-tuning,  public pretraining, and parameter freezing. Details of \ouralg are provided in \cref{algo:proposed}.

\begin{algorithm}[ht]
    \DontPrintSemicolon
    \SetKwInput{Input}{Input}
    \SetAlgoLined
    \LinesNumbered
    \Input{\serveropt with learning rate $\alpha$;\newline \clientopt with learning rate $\beta_1$ and $\beta_2$;\newline 
    clip norm $\gamma$ and noise multiplier $\sigma$;\newline (optional) pretrained model $\theta^{(0)}$}
     \For{{\it \bf round} $t = 0,1,\dots,T-1$ }{
      Sample a subset of users $\userset^{(t)}$\;
      Partition users $\userset^{(t)}$ to virtual clients $\vcset^{(t)}$ \;
      \For{{\it \bf each virtual client} $V \in \vcset^{(t)}$ {\it \bf in parallel}}{
        Initialize backbone $\theta_V^{(t,0)}=\theta^{(t)}$ \;
        Randomly initialize head $\omega_V^{(t,0)}$ \;
        \For {$k =0,\dots, K-1$}{
            Sample minibatch $B \subset \bigcup_{i \in V} \data_i $ \;
            Compute gradients $\nabla_{\theta_V} \ell_B, \nabla_{\omega_V} \ell_B$, where $\ell_B = \mathbb{E}_{(x,y)\in B} \, \ell(\langle \omega_V, \, f(\theta_V, x)\rangle, \, y) $  \;
            Update $\theta_V^{(t,k+1)}$ by $\clientopt$, $\theta_V^{(t,k)}$, $\nabla_{\theta_V} \ell_B$, $\beta_1$\;
            Update $\omega_V^{(t,k+1)}$ by $\clientopt$, $\omega_V^{(t,k)}$, $\nabla_{\omega_V} \ell_B$, $\beta_2$\;
        }
        Compute clipped model update $\Delta_V^{(t)} = \text{Clip}(\theta_V^{(t, K)} - \theta_V^{(t,0)}, \, \gamma) $\;
      }
      Aggregate model updates $\Delta^{(t)} = \nicefrac{\text{\small AddNoise} \left( \sum_{V \in \vcset^{(t)}}  \Delta_V^{(t)}, \, \sigma\gamma \right)}{|\vcset^{(t)}|}$ \;
      Update global backbone parameters $\theta^{(t+1)} = \serveropt(\theta^{(t)}, \Delta^{(t)}, \, \alpha)$ \;
     }
     \caption{ \ouralg: learning embedding model $\theta$ with user-level DP}
     \label{algo:proposed}
\end{algorithm}

\textbf{Virtual clients.}
Data heterogeneity is one of the key problems in federated optimization \citep{wang2021fieldguide}. When we train embedding models in the multi-class framework, the class space can be very large and each user may only observe a limited number of classes. In the extreme case, when training embedding models on facial images \citep{taigman2014deepface,schroff2015facenet}, each user may only have images for their own identity. This significantly limits the advantage of local updates due to client drift \citep{karimireddy2020scaffold}, and even with specialized regularization like \citep{yu2020federated}, \fedsgd \citep{mcmahan2017fedavg} with frequent aggregation and model release has to be used instead of \fedavg. It is challenging to use some specialized techniques for handling data heterogeneity \citep{wang2021fieldguide} in DP training. Instead, we propose a simple yet effective approach: randomly groups the data of sampled users into \emph{virtual clients}. 

Unlike the cross-device FL setting where the on-device data of users cannot be directly communicated, virtual clients are feasible for user data in the datacenter. It is important to guarantee that a user will not be included in two virtual clients in a single round for user-level DP, analogous to microbatches for DP-SGD and example-level DP \cite{abadi2016deep,mcmahan2018general}. When the grouping of users for virtual clients is fixed in advance across rounds, the granularity of the DP definition can slightly change: the adjacent dataset for DP is based on virtual clients (a group of users) instead of a single user, which has conceptually stronger privacy guarantees. However, when virtual clients are randomly regrouped across rounds as in \cref{algo:proposed}, we can only show user-level DP as discussed in \cref{app:vc_dp}. Virtual clients also control the interpolation between federated training and centralized training: when all users are grouped into a single virtual client, federated training is equivalent to centralized training, which removes heterogeneity but is challenging for DP mechanism. Virtual clients are used for both baseline DP-\fedavg and the proposed \ouralg method.

\textbf{Partial aggregation and local fine-tuning.}
Another challenge is the number of parameters in DP training. A common backbone $\theta$ of ResNet-50 for $128$ dimensional embedding vectors has $23.77$ million parameters. However, the parameter size of head $\omega$ can linearly grow with the number of classes. Taking FaceNet \citep{taigman2014deepface,schroff2015facenet} as an example again, $\omega$ can easily grow to $1280$ million for $10$ million identities in real-world applications. Sampled softmax \citep{jean2014using,waghmare2022efficient} can be applied to improve training efficiency. However, as both backbone $\theta$ and head $\omega$ are shared among users and need to be privatized by adding noise during training, the combined parameter size of $(\theta, \omega)$ will significantly affect the privacy utility trade-off, which cannot be mitigated by sampled softmax. 

In \ouralg, inspired by federated reconstruction \citep{singhal2021federated} and DP personalization \citep{jain2021differentially}, we only aggregate and privatize the backbone network $\theta$, which is used in inference and has fixed parameter size that does not grow with classes. A local head $\omega_V$ is randomly initialized and updated on each virtual client $V$. A fine-tuning approach is adopted for local updates, where different learning rates $\beta_1, \beta_2$ are used for the backbone $\theta_V$ and head $\omega_V$, respectively. When combined with virtual clients, the partial aggregation and local fine-tuning approach can be interpreted in various ways: each virtual client is performing transfer learning given a shared backbone network for representation learning; the data of each class are their own positive samples as well as negative samples for other classes on the same virtual client; the size of local head $\omega_V$ is also significantly smaller than $\omega$ for all classes, which is effectively a user-based sampling for softmax.

\textbf{Public pretraining.} The parameter size of the backbone to be privatized can still be large after applying partial aggregation and local fine-tuning with virtual clients, \eg, $23.77$ million for ResNet-50. Inspired by recent research on applying DP-SGD for example-level DP on large language modeling \citep{li2021large,yu2021differentially} and image classification \citep{de2022unlocking,kurakin2022toward}, we use a model pretrained on public images to initialize the DP training of the backbone network. There is a relatively clear distinction between the public and private domains for our task: we use public images collected from open webpages for pretraining, and then privately train on users' data collected in a datacenter.

\textbf{Parameter freezing.} 
Neural networks are known to be overparameterized, and not all weights are equally important \citep{zhang2019layers,frankle2021training}. Freezing some parameters to be non-trainable has been shown to be effective when the privacy budget is small \citep{sidahmed2021efficient}, especially when combined with public pretraining for large models \citep{yu2021differentially,de2022unlocking}. For backbone convolutional neural networks with normalization layers, we experiment with training parameters with all normalization layers, and some of the convolutional kernels. However, freezing is found to be less efficient in our setting that performs representation learning, instead of classification, for a moderate size model in the high-utility-moderate-noise regime.

\textbf{DP mechanism and hyperparameters.} Similar to generalized DP-\fedavg, we perform clipping for model deltas and add noise for aggregated updates. The clip norm $\gamma$ is estimated by adaptive clipping \citep{andrew2019differentially} in the parameter tuning stage. For \ouralg, we perform extensive studies on several configurations in \cref{sec:exp2}. Differentially private hyperparameter tuning \citep{papernot2021hyperparameter} is a topic out of the scope of this paper, and automating hyperparameter tuning is an important future work. Either independent Gaussian noise like DP-SGD \citep{mcmahan18learning} or tree-based correlated noise like DP-FTRL \citep{kairouz21practical} can be added. Under the same noise multiplier, \ouralg will achieve the same privacy bound as DP-\fedavg with virtual clients, while utility can be improved for training a backbone network with a large head.

%% file: dp_fedavg_alg.tex
\begin{algorithm}[thb] 
    \DontPrintSemicolon
    \SetKwInput{Input}{Input}
    \SetAlgoLined
    \LinesNumbered
    \Input{\serveropt with learning rate $\alpha$;\newline \clientopt with learning rate $\beta$;\newline clip norm $\gamma$ and noise multiplier $\sigma$;\newline (optional) pretrained model $\theta^{(0)}$}
     \For{{\it \bf round} $t = 0,1,\dots,T-1$ }{
      Sample a subset $\userset^{(t)}$ of users \;
      \For{{\it \bf each client} $i \in \userset^{(t)}$ {\it \bf in parallel}}{
        Initialize parameters $(\theta_i, \omega_i)^{(t,0)}=(\theta, \omega)^{(t)}$ \;
        \For {$k =0,\dots, K-1$}{
            Sample minibatch $B \subset \data_i $ \;
            Compute gradients $\nabla_{(\theta_i, \omega_i)} \ell_B$, where $\ell_B = \mathbb{E}_{(x,y)\in B} \, \ell(\langle \omega_i, \, f(\theta_i, x)\rangle, \, y) $ \; 
            Update $(\theta_i,\, \omega_i)^{(t,k+1)}$ by $\clientopt$, $(\theta_i, \, \omega_i)^{(t,k)}$, $\nabla_{(\theta_i, \omega_i)} \ell_B$, $\beta$\;
        }
        Compute clipped model update $\Delta_i^{(t)} = \text{Clip}((\theta_i, \omega_i)^{(t, K)} - (\theta_i, \omega_i)^{(t,0)}, \, \gamma) $\;
      }
      Aggregate model updates $\Delta^{(t)} = \nicefrac{ \text{\small AddNoise}\left( \sum_{i \in \userset^{(t)}}  \Delta_i^{(t)}, \, \sigma\gamma \right) }{ |\userset^{(t)}|}$ \;
      Update global parameters $(\theta, \omega)^{(t+1)} = \serveropt((\theta, \omega)^{(t)}, \Delta^{(t)}, \, \alpha)$ \;
     } 
     \caption{Learning embedding model $\theta$ with generalized DP-\fedavg \citep{mcmahan18learning,wang2021fieldguide}.} \label{algo:dp-fedavg}
\end{algorithm}

%% file: sec_exp.tex
\section{Experiments} \label{sec:exp2}

We conduct experiments to train image-to-embedding backbone networks with user-level DP. We use the DigiFace dataset \citep{bae2023digiface1m} of synthetic faces based on ethical and responsible development considerations, and verified that the conclusions on DigiFace are very similar to conclusions generated from experiments on natural facial images. 
We randomly split the DigiFace dataset of $110K$ identities and $1.22M$ into subsets of $98.96K$ identities with $1.10M$ images for training, $5443$ identities with $58.24K$ images for validation, and $5598$ identities with $60.82K$ images for testing. We extensively use a smaller training set, DigiFace10K, which contains the $9047$ training identities of 72 images sampled from the DigiFace training set. We also run experiments on public datasets of natural images: EMNIST, Google Landmarks Dataset (GLD)  and iNaturalist (iNat) dataset. These datasets are summarized in \cref{tab:datasets2}.

In our setting, each user holds only the images of their own identities, i.e., user-level DP is also identity-level DP. We use ResNet-50 \citep{he2016deep} and MobileNetV2 \citep{sandler2018mobilenetv2,hsu2020federated,wang2021fieldguide} as backbone networks, replace batch normalization \citep{ioffe2015batch} with group normalization \citep{wu2018group}, and use a multi-class framework with a large softmax head to train the backbone. The dimension of the embeddings are $128$ for all experiments. 

We evaluate the performance of the backbone network based on predicting identity matches from the distance between two image embeddings. By varying a threshold on the pairwise similarity, a recall versus false accept rate (FAR) curve on the test data can be generated for a trained model. A scalar value of \ourmetric is often reported. The privacy guarantees are computed by either using Renyi differential privacy (RDP) \citep{mironov2017renyi} and converting to $(\epsilon, \delta)$-DP by \citep{canonne2020discrete}, or DP-FTRL accounting without restart \citep{kairouz21practical}. More discussion on privacy accounting can be found in \cref{app:vc_dp}. We aim for single-digit $\epsilon$ when $\delta$ is small, and sometimes relax to $\epsilon \sim 20$ as we use the stronger substitute-one DP definition \citep{ponomareva2023dp}. 

We compare the proposed \ouralg with non-private oracle performance of centralized training, and baseline methods DP-\fedavg. We tune the learning rate with learning rate scheduling for standard centralized training. The centralized baseline is provided as an oracle for non-private training performance. We exclude tricks like data augmentation for either centralized or federated training as the goal is not achieving state-of-the-art performance. Virtual clients are used to improve DP-\fedavg performance, and the same tuning strategy is applied for \ouralg and DP-\fedavg. In most of the experiments, unless otherwise specified, 
we fix the hyperparameters in the federated setting and only tune the learning rates; the backbone networks are pretrained on classifying the $1000$ classes of ImageNet \citep{russakovsky2015imagenet}; both \clientopt and \serveropt are SGD optimizers with momentum $0.9$; and more details are provided in \cref{sec:exp_abl2}. Code is released at \url{https://github.com/google-research/federated/tree/master/dp_visual_embeddings}.

\ifthenelse{\boolean{cvpr}}{}{\input{exp_table_dataset}}

\ifthenelse{\boolean{cvpr}}{}{\input{app_remark_account}}

\begin{figure*}[bth]
\centering
\begin{subfigure}[b]{0.235\textwidth}
\centering
\includegraphics[width=\textwidth]{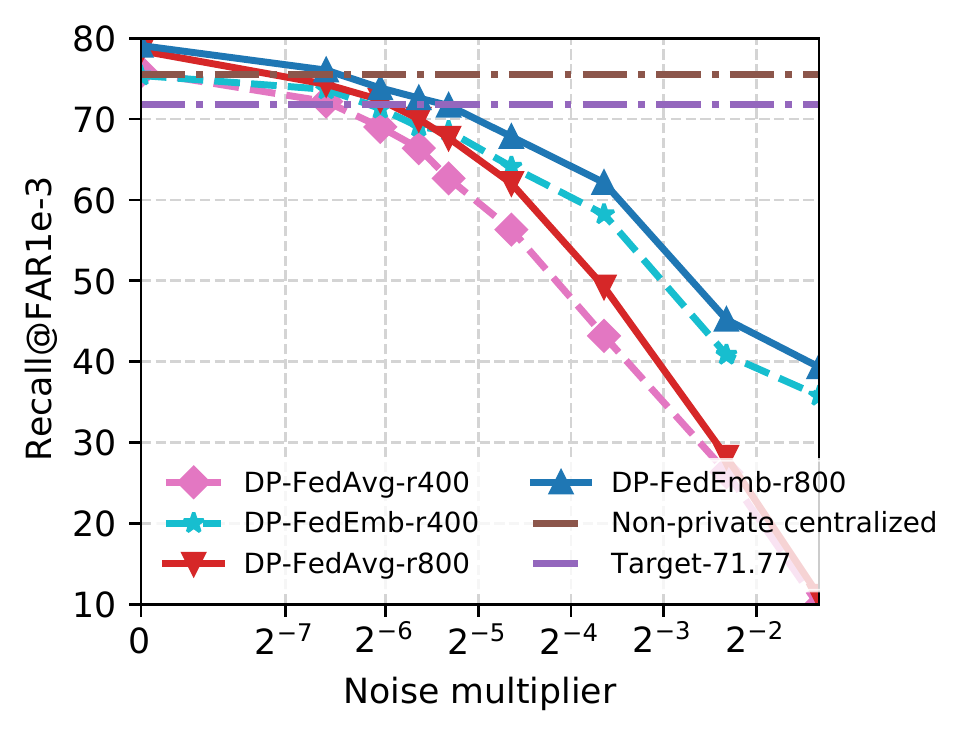}
\caption{}
\label{fig:digi-noise-util}
\end{subfigure}
\begin{subfigure}[b]{0.215\textwidth}
\centering
\includegraphics[width=\textwidth]{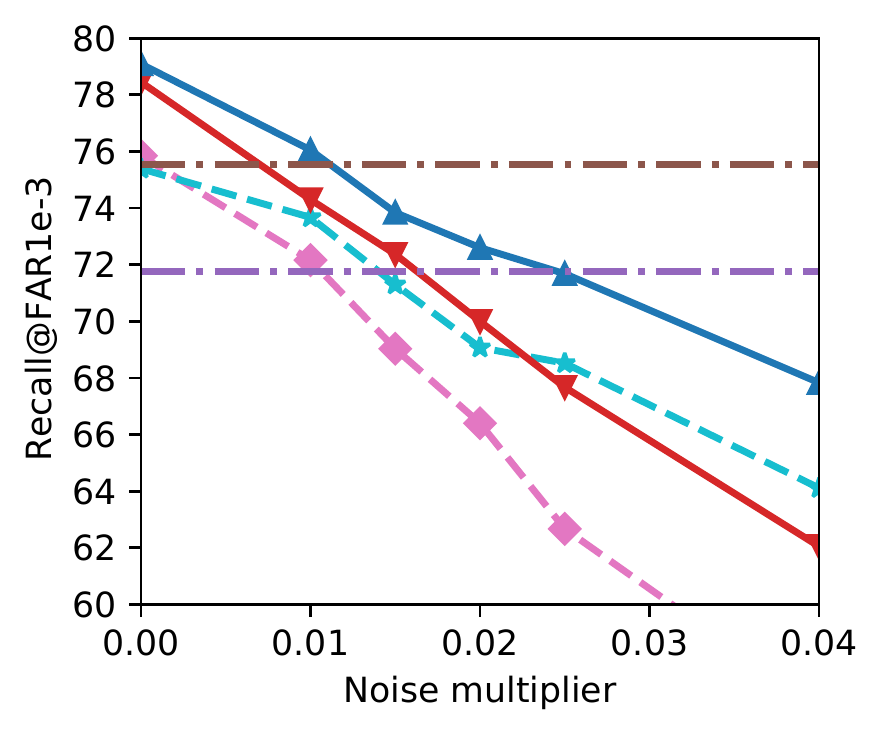}
\caption{}
\label{fig:digi-noise-util-zoomin}
\end{subfigure}
\begin{subfigure}[b]{0.255\textwidth}
\centering
\includegraphics[width=\textwidth]{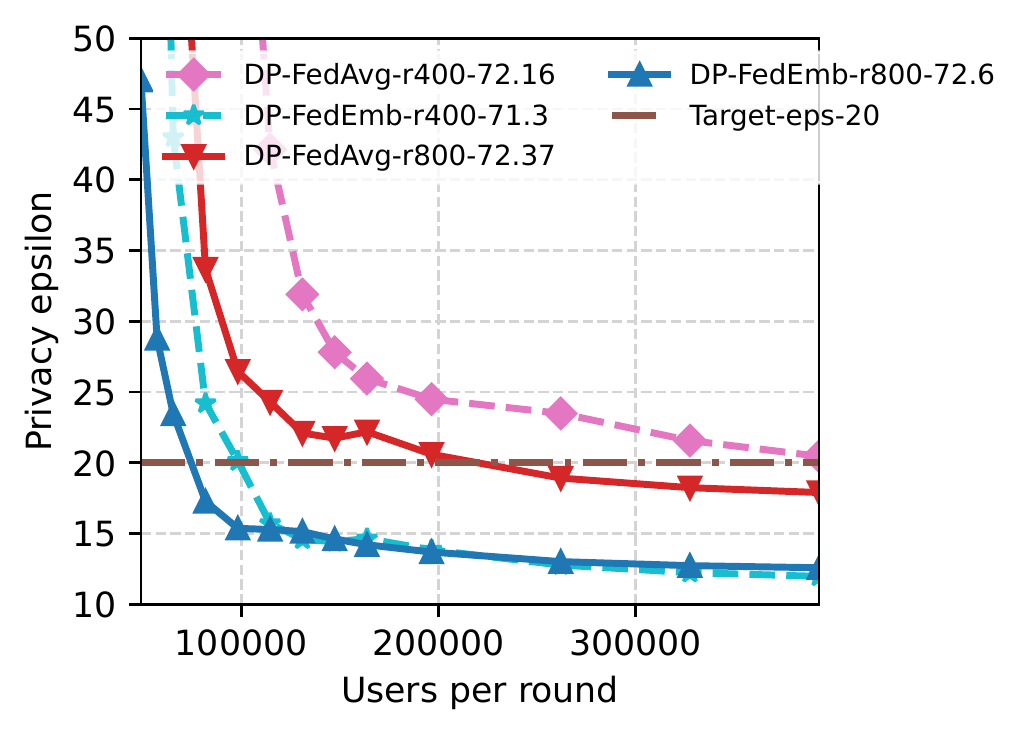}
\caption{}
\label{fig:digi-priv-comp-3m}
\end{subfigure}
\begin{subfigure}[b]{0.255\textwidth}
\centering
\includegraphics[width=\textwidth]{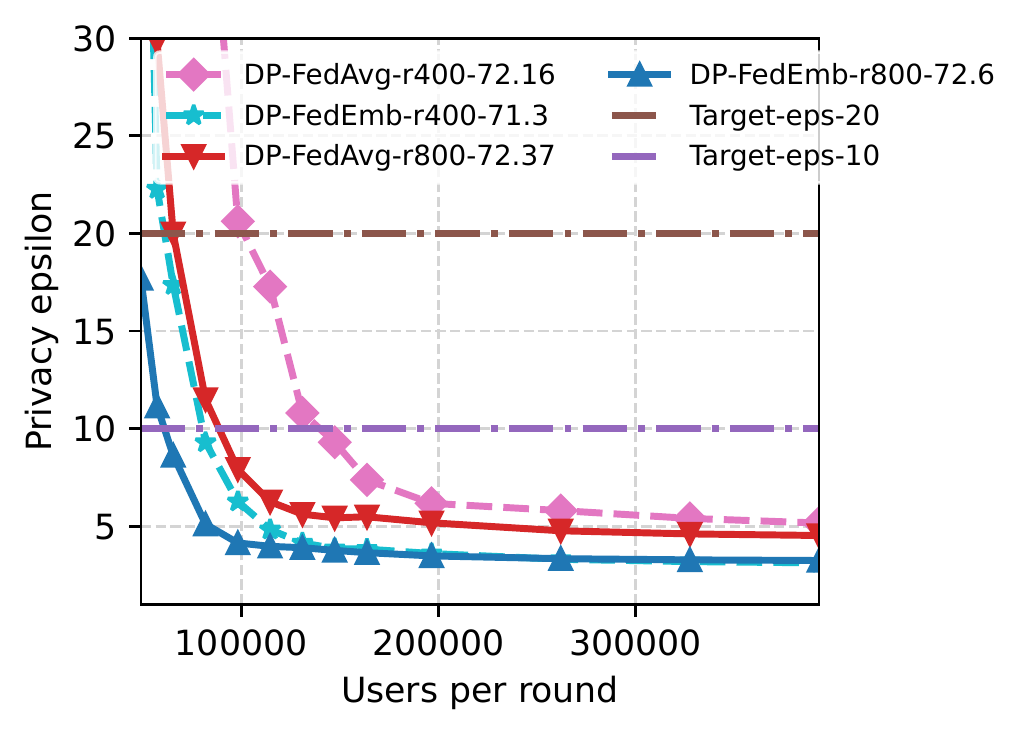}
\caption{}
\label{fig:digi-priv-comp-10m}
\end{subfigure}
\caption{(a) \Ourmetric on DigiFace validation set under different noise multiplier; (b) zoom in the high utility regime in (a);
(c) and (d) privacy-computation trade-off by extrapolating based on 3M and 10M total users, respectively. "r400" and "r800" represent the result at 400 or 800 training rounds; target 71.77\% is 95\% of centralized non-private \ourmetric at 75.55\%.
}
\label{fig:digi-dpsgd}
\ifthenelse{\boolean{cvpr}}{\vspace{-0.3cm}}{}
\end{figure*}

\ifthenelse{\boolean{cvpr}}{}{\input{exp_ftrl_curves}}

\ifthenelse{\boolean{cvpr}}{\input{exp_roc_curves_small}}{\input{exp_roc_curves_main}}

\subsection{Privacy-utility-computation trade-off} \label{sec:exp_priv_trade2}

We study the privacy-utility-computation trade-offs of training ResNet-50 on DigiFace10K in \cref{fig:digi-dpsgd} and \cref{fig:digi-dpftrl}. \Cref{fig:digi-noise-util} shows the privacy-utility trade-off. Without adding noise, the federated algorithms (DP-)FedAvg and (DP-)FedEmb can achieve even better results than the non-private centralized baseline, which is consistent with recent empirical and theoretical justifications that FedAvg is more accurate when learning representations \cite{collins2022maml,collins2022fedavg}. 
When the same noise multiplier is used, \ie, under same privacy budget, \ouralg outperforms DP-\fedavg; and the margin increases when increasing the noise. We can observe the advantage of \ouralg over DP-\fedavg even if we only have a small number of identities in DigiFace: the size of the head is $9047\times128 \sim1.16M$, only 4.6\% of the backbone ResNet-50 with $23.77M$ parameters. 
For large scale data with 10 million identities, the size of the head can grow to $1280M$, which is much larger than the backbone networks, and DP-\fedavg can easily fail in such settings.

It can be difficult to achieve a strong formal differential privacy bound without significantly hurting utility for DigiFace10K, which only has a small number of total users. We consider the practical setting of more available users, and study the privacy-computation trade-off in \cref{fig:digi-priv-comp-3m,fig:digi-priv-comp-10m,fig:digi-ftrl-eps,fig:digi-ftrl-zcdp} based on extrapolation. 
A key hypothesis following \citep{mcmahan18learning,kairouz21practical} is used: utility (Recall@FAR) is non-decreasing when simultaneously increasing the number of clients per round and noise multiplier. The hypothesis is based on the fact that the signal-to-noise ratio is non-decreasing when linearly increasing the number of clients per round and noise multiplier, and has been verified in practice \citep{ramaswamy2020training,dpftrl_blogpost}. 

We first choose the noise multiplier that is within $5\%$ of \ourmetric compared to centralized non-private training in \cref{fig:digi-noise-util,fig:digi-ftrl-noise-util}, based on simulation that samples $64$ virtual clients that each have $32$ users per round: \ourmetric is $72.16$ for running DP-\fedavg with $0.01$ noise multiplier for $400$ rounds, $72.37$ for running DP-\fedavg with $0.015$ noise multiplier for $800$ rounds, $71.3$ for running \ouralg with $0.015$ noise multiplier for $400$ rounds,  $72.6$ for running \ouralg with $0.02$ noise multiplier for $800$ rounds, and $71.85$ for running DP-FTRL-FedEmb with $0.26$ noise multiplier for $800$ rounds. Then we linearly increase the number of users sampled and use the increased noise multiplier in RDP accounting to compute privacy bound $\epsilon$ given $\delta=10^{-7}$ to generate \cref{fig:digi-priv-comp-3m,fig:digi-priv-comp-10m,fig:digi-ftrl-eps}, and compute zCDP for \cref{fig:digi-ftrl-zcdp}. Comparing curves of r400 and r800,  training longer with larger noise is more effective than training shorter with smaller noise. \Cref{fig:digi-priv-comp-3m} suggests ${\sim}98K$ users per round is enough for \ouralg to achieve single-digit $\epsilon$ if $3M$ users are available, while ${\sim}57K$ users per round are needed if $10M$ users are available in \cref{fig:digi-priv-comp-10m}. $98K$ users is $48\times$ the number of users per round in our current simulation, which can be achieved by training with $8\times$ computing resources for $6\times$ longer. In \cref{fig:digi-ftrl-eps}, there is a crossover point when using DP-FTRL versus DP-SGD for \ouralg, and DP-FTRL is more effective for relatively large privacy $\epsilon$. \Cref{fig:digi-ftrl-zcdp} shows that DP-FTRL-FedEmb can achieve zCDP smaller than $2.6$, as used by US Census Bureau \citep{bureau2021disclosure}, when $8\times$ users per round and $10M$ total users are available. 

\begin{table*}[ht]
    \centering
    \begin{tabular}{|c|c|c|c|c|c|c|}
    \hline
    \multirow{2}{*}{Algorithm} & \multicolumn{2}{c|}{Hyperparameters} & \multicolumn{2}{c|}{Privacy (10M users)} & \multicolumn{2}{c|}{\Ourmetric} \\ 
    \cline{2-7} 
    & Noise & \text{SerLR} & RDP-$\epsilon$  & zCDP & Validation  & Test   \\ 
    \hline
    Centralized & $0$ & $0.05$ & $\infty$ & $\infty$ & $75.55\pm0.05$ & $75.53\pm0.12$   \\
    \hline
    DP-FedAvg &  $0.015\times64$ & $0.5$ & $5.62$ & - & $72.57\pm0.12$ & $72.37\pm0.09$  \\
    \hline
    DP-FedEmb & $0.02\times64$ & $0.2$ & $3.90$ & - & $72.63\pm0.05$  & $72.37\pm0.09$ \\
    \hline
    {\footnotesize DP-FTRL-FedEmb} & $0.26\times64$ & $0.2$ & $9.67$ & $1.28$ & $72.2\pm0.29$  & $71.87\pm0.26$ \\
    \hline
    \end{tabular}
    \caption{Quantitative results of privacy and utility on the DigiFace10K dataset. The client learning rate and clip norm for federated algorithms are $0.002$ and $0.6$, respectively; $\delta=10^{-7}$ for privacy guarantees. Centralized training has a standard learning rate scheduling, while tricks like data augmentation are excluded for all methods. The privacy guarantees are extrapolated based on $10M$ users and $131K$ users per round. A strong privacy guarantee can be achieved within a $5\%$ drop on \ourmetric.} 
    \label{tab:digi-results}
\end{table*}

\subsection{Model evaluation} \label{sec:exp_model_eval2}
In \cref{tab:digi-results}, we summarize the quantitative results from the privacy-utility-computation trade-off analysis in \cref{sec:exp_priv_trade2}. Each experiment runs three times to compute the mean and standard deviation. For similar \ourmetric on the DigiFace validation set, \ouralg achieves stronger privacy guarantee than baseline DP-\fedavg, and the advantage of \ouralg is expected to be more pronounced if a head for larger $10M$ identities is used for training. When $10M$ users are available and $64\times$ users per round in training, privacy $\epsilon=3.90$ of single digit and zCDP$=1.28$ smaller than $2.6$ can be achieved when \ourmetric is within a $5\%$ drop compared with non-private centralized training. In addition to validation performance, the private models also perform well on the left-out test dataset. 
\Cref{fig:curve-roc2} presents the training curves and ROC curves for comparing \ouralg and DP-\fedavg under the same privacy budget. \ouralg outperforms DP-\fedavg in all training rounds, and trains a stronger private model with better recall at different false accept rates.

\subsection{Ablation study} \label{sec:exp_abl2}
We primarily use MobileNetV2 for ablation studies on DigiFace10K for two reasons: MobileNetV2 is smaller and faster for training in experiments; to test the generalization of \ouralg and avoid overfitting on ResNet-50.

\ifthenelse{\boolean{cvpr}}{\input{exp_param_freeze_small}}{\input{exp_param_freeze}}

\ifthenelse{\boolean{cvpr}}{}{\input{exp_ablation_curves}}

\begin{figure*}[thb]
\centering
\begin{subfigure}[b]{0.215\textwidth}
\centering
\includegraphics[width=\textwidth]{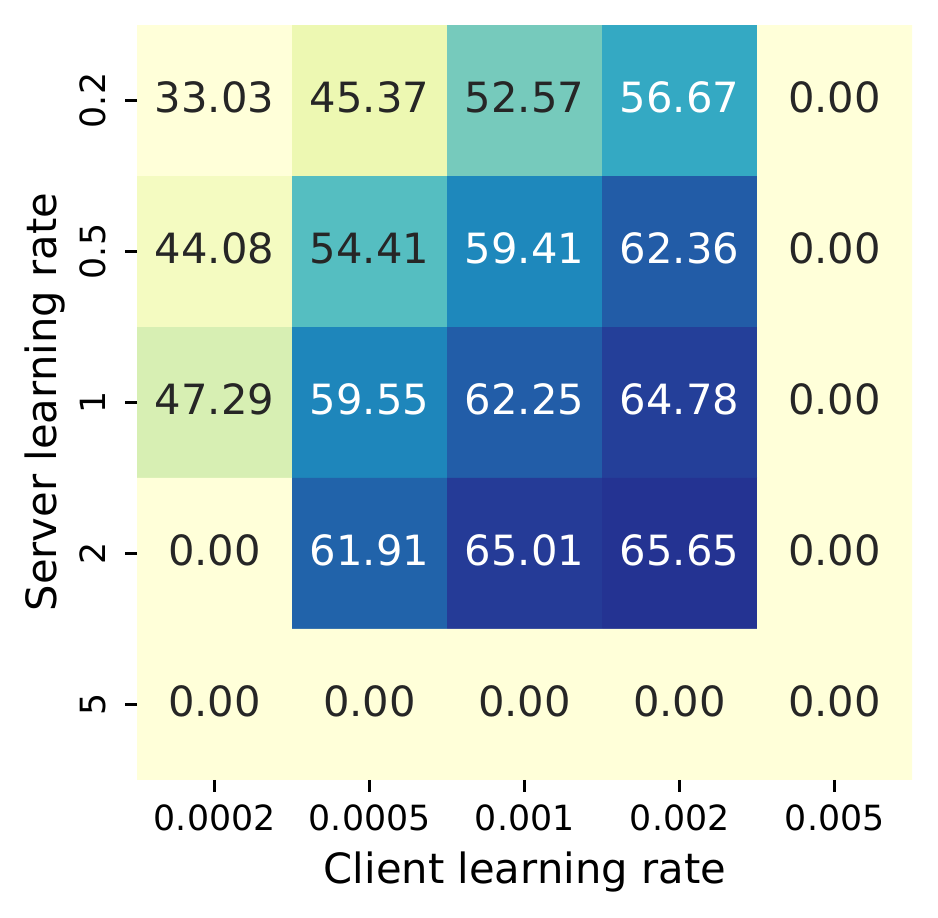}
\caption{}
\label{fig:digi-abl-lr-map400}
\end{subfigure}
\begin{subfigure}[b]{0.25\textwidth}
\centering
\includegraphics[width=\textwidth]{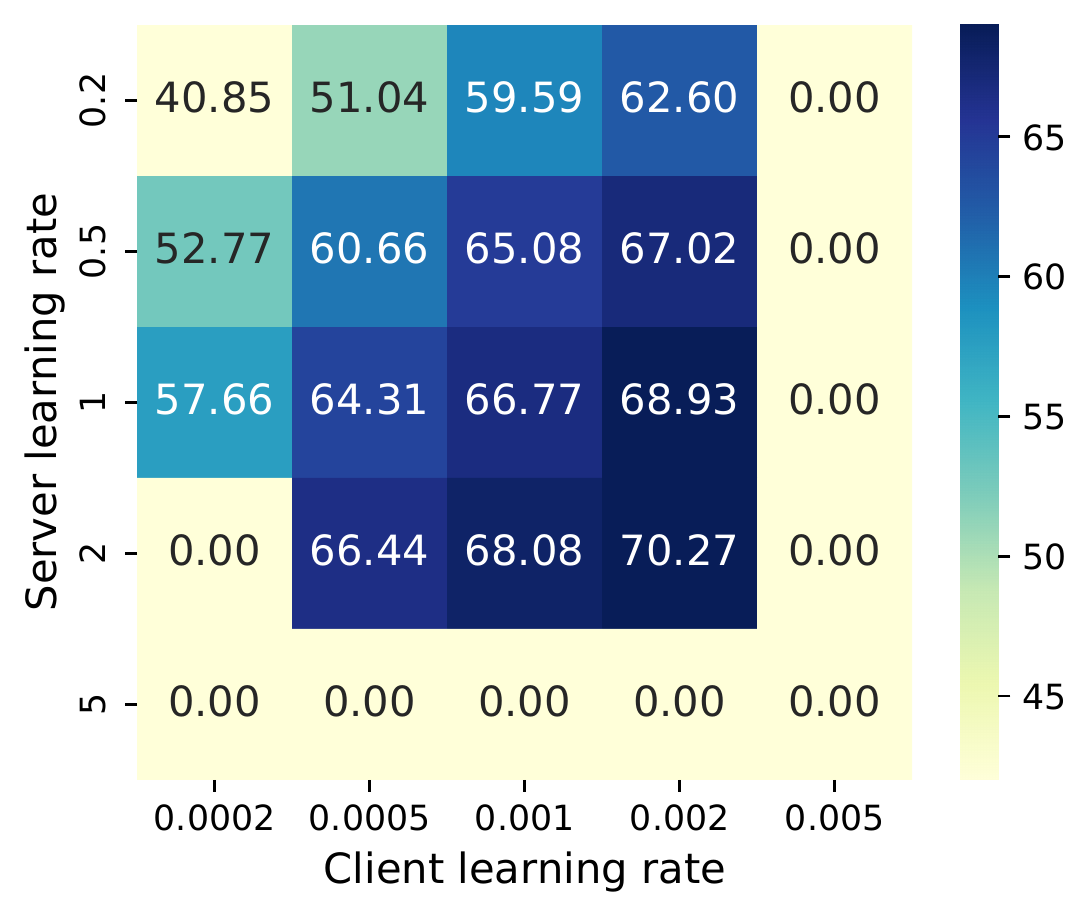}
\caption{}
\label{fig:digi-abl-lr-map800}
\end{subfigure}
\begin{subfigure}[b]{0.245\textwidth}
\centering
\includegraphics[width=\textwidth]{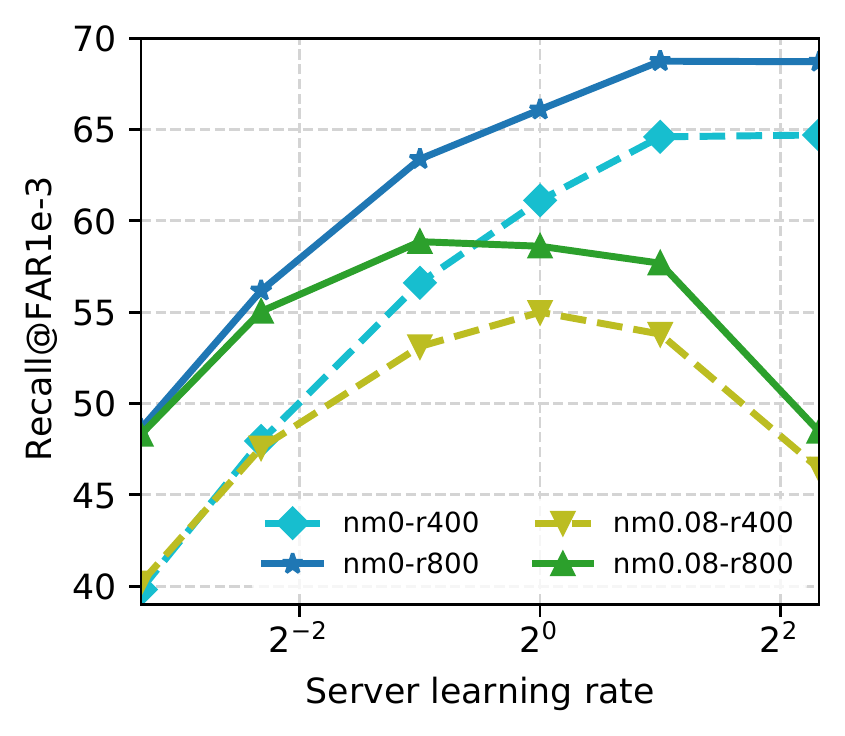}
\caption{}
\label{fig:digi-abl-slr-mobile}
\end{subfigure}
\begin{subfigure}[b]{0.26\textwidth}
\centering
\includegraphics[width=\textwidth]{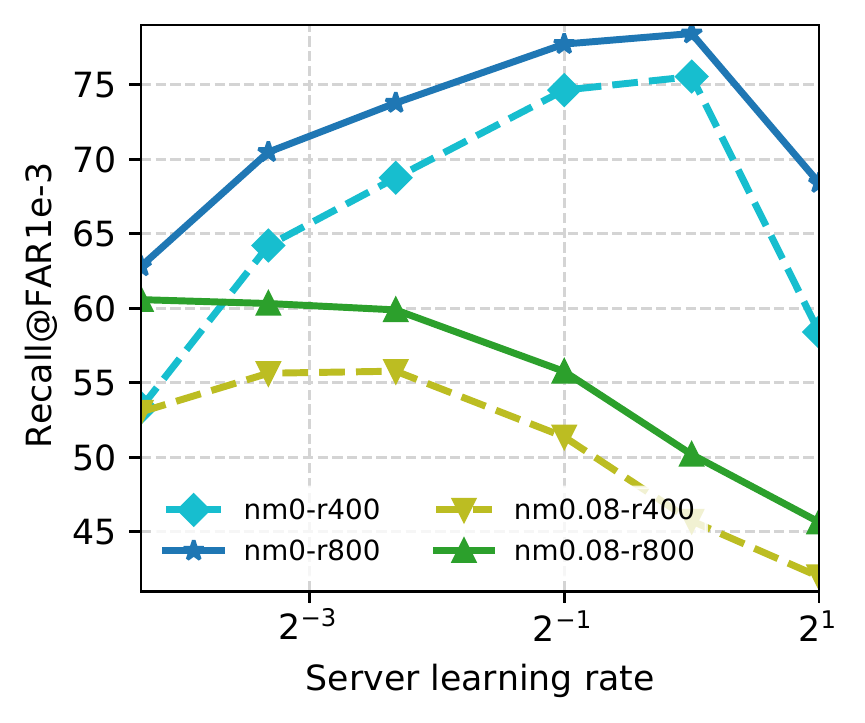}
\caption{}
\label{fig:digi-abl-slr-res}
\end{subfigure}
\caption{ Study for learning rates. (a) and (b) \ourmetric after training MobileNetV2 using \ouralg with adaptive clipping \citep{andrew2019differentially} and 0 noise for 400 rounds and 800 rounds, respectively; (c) and (d) varying server learning rate with fixed client learning rate and clip norm for MobileNetV2 and ResNet-50.}
\label{fig:ablation2_lr}
\ifthenelse{\boolean{cvpr}}{\vspace{-0.3cm}}{}
\end{figure*}

\textbf{Federated settings.} In the above experiments, we fix important hyperparameters for the federated setting: users per virtual client is $32$, virtual clients per round is $64$, examples per client is capped at $2048$, the head learning rate (LR) scale $\beta_2/\beta_1$ is $100$, the buffer size for data shuffling on clients is $2048$, and the batch size for local SGD is $32$. In \cref{fig:ablation2}, instead of tuning these hyperparameters in advance, we conduct a study on these hyperparameters to understand \ouralg.   
Among these hyperparameters, \cref{fig:digi-abl-dc,fig:digi-abl-hls} suggest virtual clients and head LR scale are particularly important for \ouralg to be on par with non-private centralized training, which is an important contribution of this work. \Cref{fig:digi-abl-dc,fig:digi-abl-cpr,fig:digi-abl-epc} suggest users per virtual client, clients per round, and examples per client only need to be large enough under privacy consideration and computation resources for the best practice. The head LR scale has a large tuning range between 50 and 500 in \cref{fig:digi-abl-hls}. \Ourmetric is not sensitive to shuffle buffer size in \cref{fig:digi-abl-csbs}. The model utility can be potentially improved if we further tune the client batch size as suggested by \cref{fig:digi-abl-cbs}. We fixed the learning rate and other hyperparameters while varying one of the hyperparameters in the ablation study. How to automate tuning, especially tuning with differential privacy guarantees, is an important future work.

\textbf{Learning rate (LR).} For experiments in \cref{sec:exp_priv_trade2,sec:exp_model_eval2}, server and client learning rates are first tuned for non-private federated training with adaptive clipping of quantile $0.5$ \citep{andrew2019differentially}. Then server learning rate is tuned when adding noise for private tuning, while estimated clip norm and client learning rate are fixed. The tuning range of learning rates are $\{1, 2, 5\}*10^{-n}$. \Cref{fig:digi-abl-lr-map400,fig:digi-abl-lr-map800} suggest the optimal learning rates are similar for training 400 rounds or 800 rounds in non-private training. We then fix the client learning rate to be $0.002$ and use the estimated clip norm $0.6$ for MobileNetV2 in \cref{fig:digi-abl-slr-mobile}, 
and observe the fixed clip results are very similar to adaptive clipping results.
The best server learning rate for private training with noise can be smaller than non-private training, where the difference is even more notable for larger model ResNet-50 in \cref{fig:digi-abl-slr-res}. 

\textbf{Variants of \ouralg.} 
We use local fine-tuning with different learning rates $\beta_1, \beta_2$ to update backbone and head parameters. An alternative is to reconstruct the head first before fine-tuning the backbone \citep{singhal2021federated,kumar2022fine}. We empirically find that head reconstruction can only achieve similar performance as the proposed fine-tuning when there are same or more number of updates on the backbone network, and hence use local fine-tuning for efficiency. It is also possible to use binary or triplet loss within a virtual clients. In our preliminary results, they achieve inferior results compared to \ouralg that uses multi-class cross-entropy loss. We leave other improvement like arcface loss \citep{deng2019arcface} as future work.

\subsection{Additional results}

\ifthenelse{\boolean{cvpr}}{\input{exp_table_add_small}}{\input{exp_table_add}}

We run additional experiments of ResNet-50 on the larger DigiFace of $98.96K$ users. Because a lot of users in DigiFace have only $5$ images, we set each virtual client to contain $64$ users, and use $3072$ samples per virtual client. We sample $128$ virtual clients per round, and a relatively large noise $1.39$ in RDP accounting can achieve $\epsilon=24.86, \delta=10^{-5}$ without extrapolation. The utility measured by \ourmetric is shown in \cref{tab:additional2}. Though the utilities of both methods are significantly degraded by the large noise, \ouralg is much better than the DP-\fedavg baseline because the noise is only added to backbone of ${\sim}2.4M$ parameters instead of backbone plus head of ${\sim}15M$ parameters. 
\ifthenelse{\boolean{cvpr}}{The full table including EMNIST~\citep{emnist} results for reproducibility and hyperparameter choices is provided in \cref{tab:additional2_2} in \cref{app:exp}.}{
\BlankLine
For EMNIST~\citep{emnist}, we train the embedding model on images of class $0-35$ and test on images of class $36-62$. Using a relatively large noise multiplier $0.62$ for $200$ rounds, and sampling $8$ users per virtual client and $32$ virtual clients per round, $\epsilon=9.28,\delta=10^{-4}$ can be achieved given $6800$ users. A small nework with two convolutional layers similar to LeNet \citep{lecun1998gradient} is used as the backbone network, and no pretrained model is used for initialization. We provide results on EMNIST primarily for reproducibility as the scale of EMNIST is smaller than the other datasets used in this draft. 
}

In \cref{tab:additional2}, we also conduct experiments with MobileNetV2 on Google Landmark Dataset (GLD) \citep{weyand2020google,hsu2020federated,gldv2} and iNaturalist (iNat) dataset \citep{liu2015faceattributes,hsu2020federated,inat}
to demonstrate the generalization of \ouralg.  We use a public model pretrained on ImageNet, and report extra approximate recall@FAR by computing pairwise similarity for minibatches, which is easy to reproduce and consistent with the all pair recall@FAR. We fix the hyperparameters for the federated settings, and compare the performance of \ouralg and DP-\fedavg under the same privacy budget (noise multiplier). Since each user already has multiple classes in GLD, we use a smaller number of users, $8$, in each virtual client. We also use a smaller number of virtual clients per round, $32$, for fast experiments and strong sampling effect. 
\Ourmetric of \ouralg and DP-\fedavg with small noise multiplier $0.02$ on GLD outperforms centralized training.
For iNat, we use an even smaller four users per virtual client and train for only $400$ rounds, and use a relatively large noise multiplier $0.5$ to get a single-digit $\epsilon=16.06$ DP guarantee for $9275$ users.  We report recall@FAR=$0.1$ instead of \ourmetric for the challenging iNat task.
In all experiments, \ouralg consistently outperforms DP-\fedavg.

%% file: exp_table_dataset.tex
\begin{table*}[ht]
    \centering
    \begin{tabular}{|c|c|c|c|c|c|c|c|c|c|}
    \hline
    \multirow{2}{*}{Dataset}  & \multicolumn{3}{c|}{Train} & \multicolumn{3}{c|}{Validation} & \multicolumn{3}{c|}{Test} \\ 
    \cline{2-10} 
    & Users & Classes & Images & Users & Classes & Images & Users & Classes & Images \\
    \hline
    DigiFace & $98.96K$ & $98.96K$  & $1.10M$ & $5443$  & $5443$  & $58.24K$ & $5598$ & $5598$ & $60.82K$ \\
    \hline
    DigiFace10K & $9047$ & $9047$  & $0.65M$ & \multicolumn{3}{c|}{-} & \multicolumn{3}{c|}{-} \\
    \hline
    EMNIST & $6800$ & $36$ & $0.58M$ & $3400$ & $26$ & $17.68K$ & \multicolumn{3}{c|}{-} \\
    \hline 
    GLD 
    & $1262$ & $2028$ & $0.18M$ & - & $2028$ & $19.53K$ & \multicolumn{3}{c|}{-} \\
    \hline
    iNat 
    & $9275$   & $1203$ & $0.12M$  & -  & $1203$ & $35.64K$  & \multicolumn{3}{c|}{-} \\
    \hline
    \end{tabular}
    \caption{The statistics of simulation datasets. The Google Landmarks Dataset (GLD)  and iNaturalist (iNat) dataset are preprocessed by Tensorflow Federated \citep{gldv2,inat}. For the training of EMNIST, we use images of class $0-35$ in the union of the $3400$ train and test clients in Tensorflow Federated dataset \citep{emnist}; and use images of class $36-62$ in the $3400$ test clients for validation. The shape of an image is $112\times112$ for DigiFace/DigiFace10K, $224\times224$ for GLD and iNat, and $28\times28$ for EMNIST. 
    } 
    \label{tab:datasets2}
\end{table*}

%% file: app_remark_account.tex
\ifthenelse{\boolean{cvpr}}{
\section{Remark on privacy accounting}
}{
\subsection{Remark on privacy accounting}
}\label{app:vc_dp}

The accounting and differential privacy definition of \ouralg depends on the DP mechanism applied in noise addition. If independent Gaussian noise similar to DP-SGD \citep{abadi2016deep,mcmahan18learning} is used in \cref{algo:proposed}, we adopt the substitute-one notation for DP definition \citep{dwork2014algorithmic,vadhan2017complexity} and leverage privacy amplification for uniform sampling. If tree-based noise similar to DP-FTRL \citep{kairouz21practical} is used in \cref{algo:proposed}, we adopt the add-or-remove with special element notation in \citep{kairouz21practical} for DP definition. We  uniformly sample users in each round of \cref{algo:dp-fedavg,algo:proposed} in simulation. Though DP-FTRL~\citep{kairouz21practical} assumes a different data streaming pattern, the practical effect is likely negligible. We use implementation in \citep{pldlib} for RDP accounting for \ouralg, and use the open-sourced implementation by \citep{kairouz21practical} for DP-FTRL-FedEmb. While privacy loss distribution (PLD) \citep{koskela2020tight,doroshenko2022connect} accounting can be tighter than RDP accounting, the current implementation~\citep{pldlib} does not support substitute-one and uniform sampling. Future improvement on privacy accounting can help further improve the guarantees obtained in our experiments. 

\textbf{Accounting for virtual clients.}
We provide more discussion on the accounting of virtual clients proposed in this paper. 
We consider user-level DP where datasets adjacency in the DP definition is based on changing all data of a single user, which is one kind of group-level DP stronger than example-level DP. 
Under virtual clients described in \cref{algo:proposed} and \cref{sec:dp-fedemb}, though we cannot formally show the (stronger) "virtual client"-level DP due to the randomized grouping of users, we can show user-level DP by the following key idea: when one user is replaced in one round, at most one virtual client is affected; the sensitivity is controlled by clipping the updates from virtual clients; noise is added proportional to clip norm, and hence proportional to sensitivity; a formal guarantee for Gaussian mechanism can be shown for noise proportional to sensitivity. The same logic can be used to prove for microbatches \citep{mcmahan2018general} in DP-SGD for example-level DP, which is analogy to virtual clients in DP-FedAvg and \ouralg for user-level DP. 

There is indeed a nuance in applying virtual clients in practice. Although add-or-remove-one neighboring relationship is popular in DP definition, it can be challenging in virtual clients. Following a worst case reasoning, adding or removing one user in a virtual client can \emph{arbitrarily} change the signal of the virtual client. Even though the norm of virtual clients is clipped, the sensitivity of the mechanism may be doubled. Trying to adopt the add-or-remove-one DP definition potentially cause the sensitivity of virtual clients (of more than one user) to double compared to without grouping users by virtual clients. Whether a tighter privacy accounting mechanism for add-or-remove-one DP can be developed without doubling sensitivity is an open problem. However, the sensitivity is consistent between virtual clients (of more than one user) and without virtual clients for the substitute-one DP definition.  

Another nuance comes from the amplification by sampling used to achieve strong privacy guarantees. For add-or-remove-one DP definition, Poisson sampling \citep{abadi2016deep} is assumed for privacy accounting but not enforced in simulation. By adopting the substitute-one DP definition, our accounting assumption and simulation consistently use uniform sampling. Conceptually, the substitute-one DP guarantees can be twice as strong as add-or-remove-one DP guarantees \citep[Section 2.1.1]{ponomareva2023dp}. Though the DP guarantees of different DP definition is not directly comparable, we relax the target DP guarantees of  $\epsilon \leq 10$ for add-or-remove-one DP to $\epsilon \leq 20$ for substitute-one DP in practice \citep[Section 5.2.2]{ponomareva2023dp}. Hence we use substitute-one DP definition in this paper~\footnote{In a previous version of the draft, we use privacy accounting for add-or-remove-one DP definition but did not account for the sensitivity inflation of virtual clients. We have correct the privacy guarantees to consistently use the substitute-one DP definition, and it does not affect our conclusion. Virtual clients are used for both DP-FedEmb and DP-FedAvg, which is necessary under the extreme heterogeneity, for example, when each user only has images of a single identity. Both DP-FedEmb and DP-FedAvg can achieve user-level DP and are compared under the same DP definition.}. Note that all the nuances discussed also apply to microbatches and DP-SGD, which is overlooked in the past. 

%% file: exp_ftrl_curves.tex
\begin{figure*}[htb]
\centering
\begin{subfigure}[b]{0.35\textwidth}
\centering
\includegraphics[width=\textwidth]{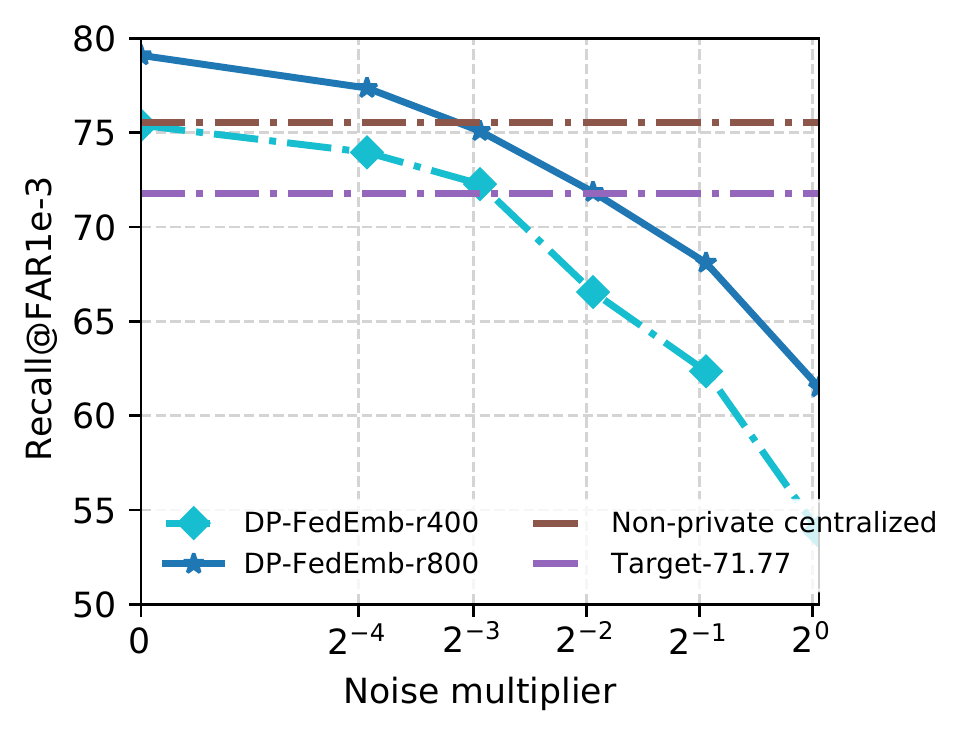}
\caption{}
\label{fig:digi-ftrl-noise-util}
\end{subfigure}
\begin{subfigure}[b]{0.33\textwidth}
\centering
\includegraphics[width=\textwidth]{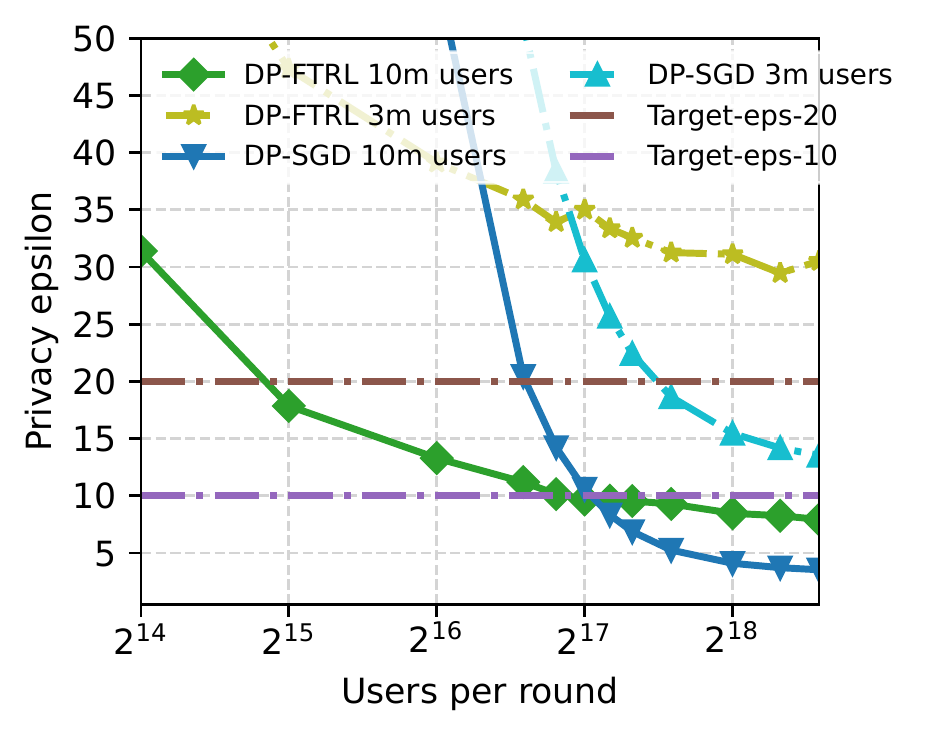}
\caption{}
\label{fig:digi-ftrl-eps}
\end{subfigure}
\begin{subfigure}[b]{0.3\textwidth}
\centering
\includegraphics[width=\textwidth]{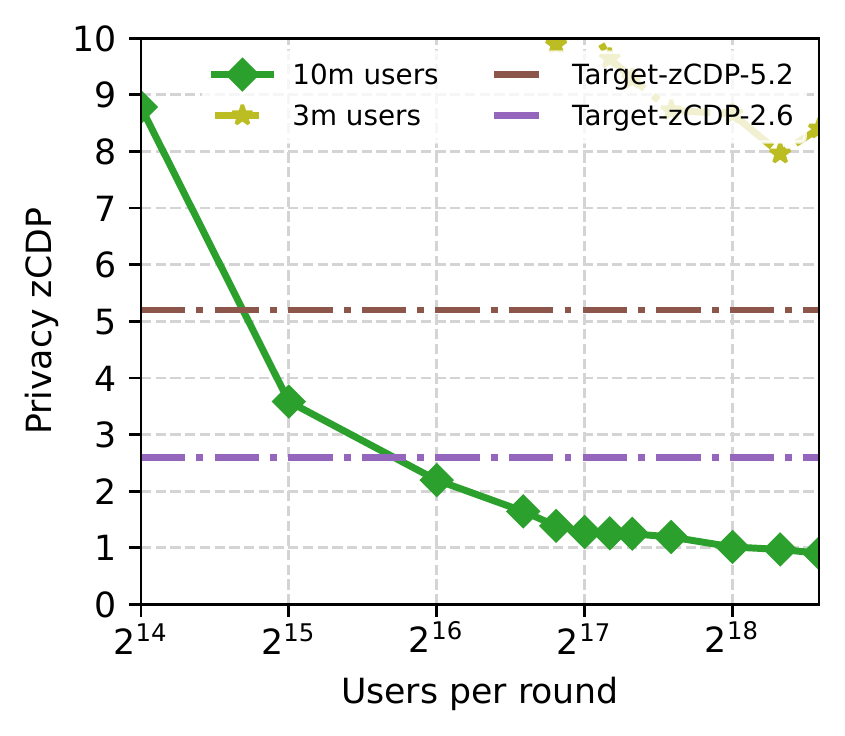}
\caption{}
\label{fig:digi-ftrl-zcdp}
\end{subfigure}
\caption{(a) \Ourmetric on DigiFace validation set under different noise multiplier when use DP-FTRL \citep{kairouz21practical} for \ouralg.
(b) and (c) privacy-computation trade-off by extrapolating based on $3M$ and $10M$ total users; $\epsilon$ by RDP accounting for \ouralg and DP-FTRL-FedEmb, and zCDP for DP-FTRL-FedEmb are reported, respectively.
}
\label{fig:digi-dpftrl}
\end{figure*}

%% file: exp_roc_curves_small.tex
\begin{figure}[htb]
\centering
\begin{subfigure}[b]{0.232\textwidth}
\centering
\includegraphics[width=\textwidth]{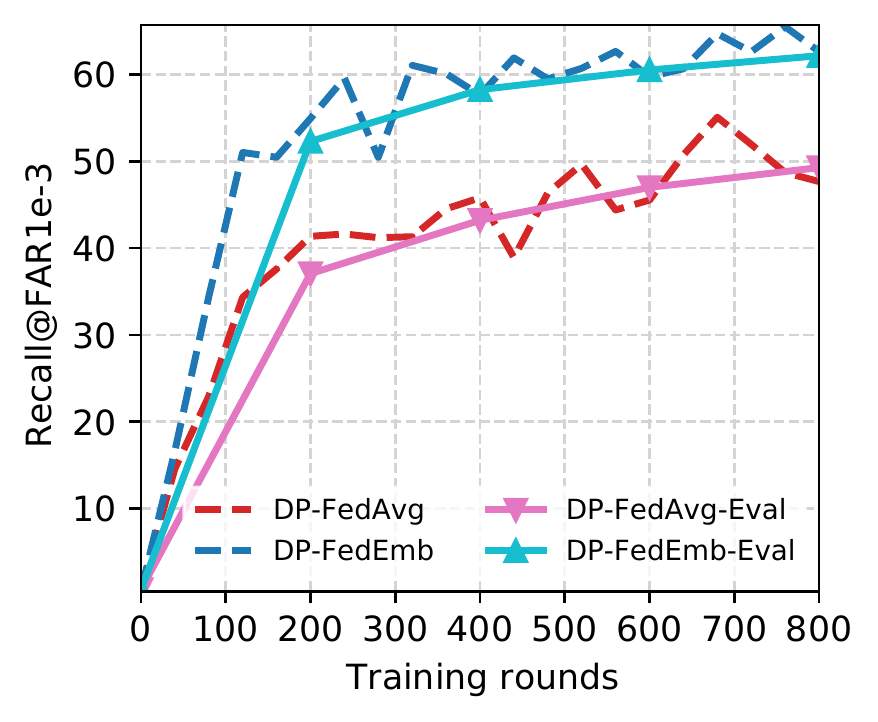}
\caption{}
\label{fig:digi-training-curve}
\end{subfigure}
\begin{subfigure}[b]{0.24\textwidth}
\centering
\includegraphics[width=\textwidth]{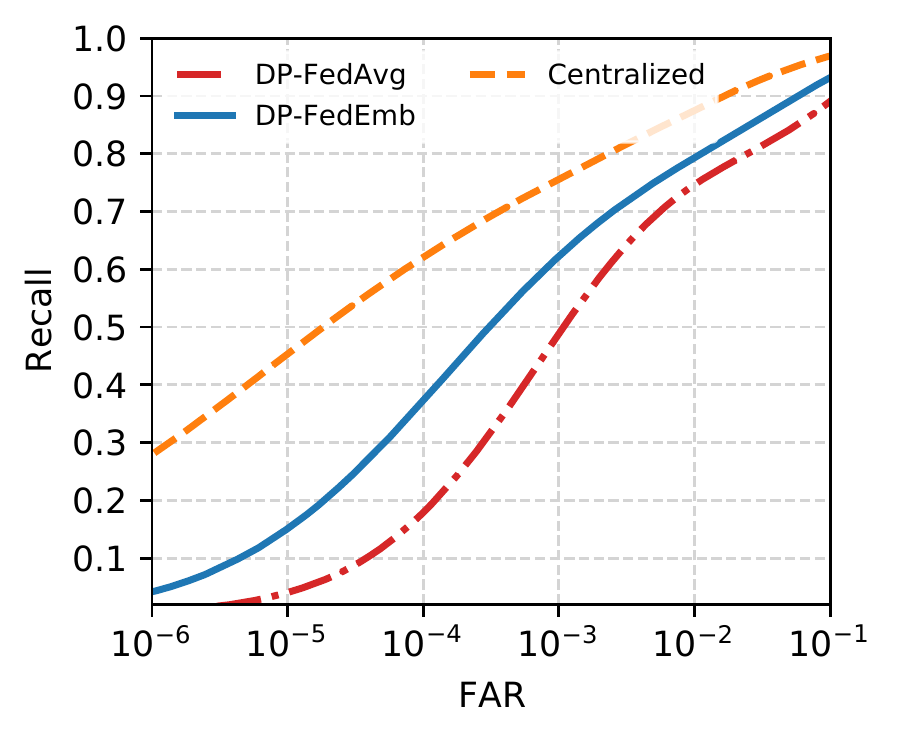}
\caption{}
\label{fig:digi-ln-roc}
\end{subfigure}
\caption{ \ouralg and DP-\fedavg on DigiFace10K under the same privacy budget with noise multiplier $0.08$, and additional oracle baseline of centralized training w/o DP: (a) \ourmetric on validation dataset during training; dashed lines are approximation by sampling a subset of validation users; (b) ROC curve of trained model on test set with log scale x-axis.
}
\label{fig:curve-roc2}
\ifthenelse{\boolean{cvpr}}{\vspace{-0.2cm}}{}
\end{figure}

%% file: exp_roc_curves_main.tex
\begin{figure*}[htb]
\centering
\begin{subfigure}[b]{0.31\textwidth}
\centering
\includegraphics[width=\textwidth]{figs2/digi-training-curve.pdf}
\caption{}
\label{fig:digi-training-curve}
\end{subfigure}
\begin{subfigure}[b]{0.32\textwidth}
\centering
\includegraphics[width=\textwidth]{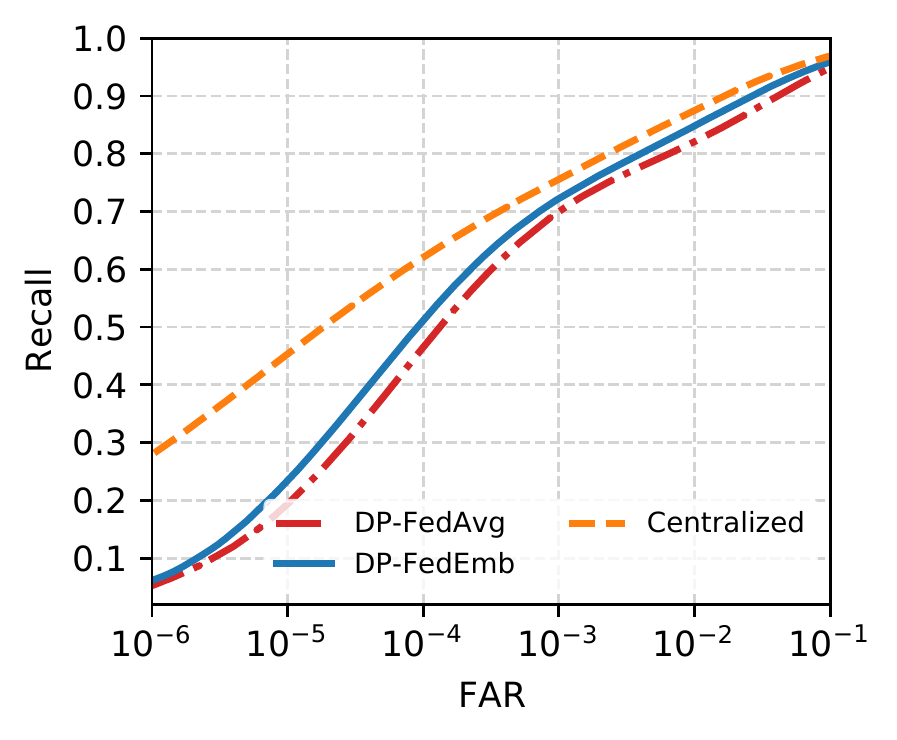}
\caption{}
\label{fig:digi-sn-roc}
\end{subfigure}
\begin{subfigure}[b]{0.32\textwidth}
\centering
\includegraphics[width=\textwidth]{figs2/test-roc-median-noise.pdf}
\caption{}
\label{fig:digi-ln-roc}
\end{subfigure}
\caption{ \ouralg and DP-\fedavg on DigiFace10K under the same privacy budget, and additional oracle baseline of centralized training w/o DP: (a) \ourmetric on validation dataset during training, and dashed lines are approximation by sampling a subset of validation users; (b) and (c) ROC curve of trained model on test set, with noise multiplier $0.02$ and $0.08$, respectively; the x-axis in (b) and (c) are in logarithmic scale.
}
\label{fig:curve-roc2}
\end{figure*}

%% file: exp_param_freeze_small.tex
\begin{figure}[tb]
\centering
\begin{subfigure}[t]{0.255\textwidth}
\centering
\includegraphics[width=\textwidth]{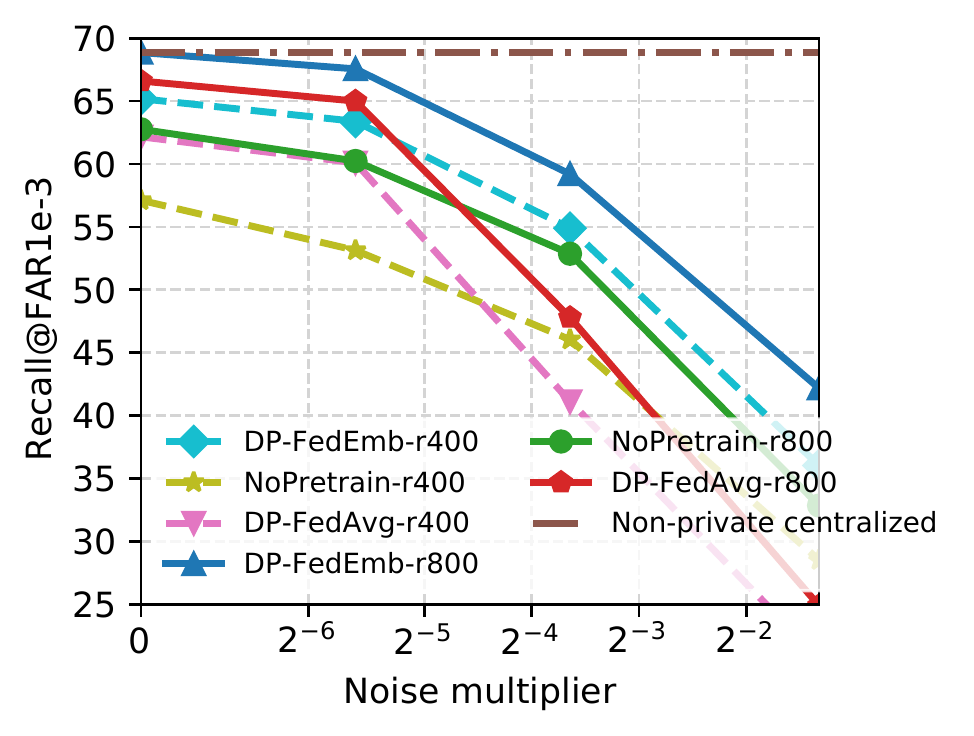}
\caption{}
\label{fig:digi-mobilenet-noise}
\end{subfigure}
\begin{subfigure}[t]{0.215\textwidth}
\centering
\includegraphics[width=\textwidth]{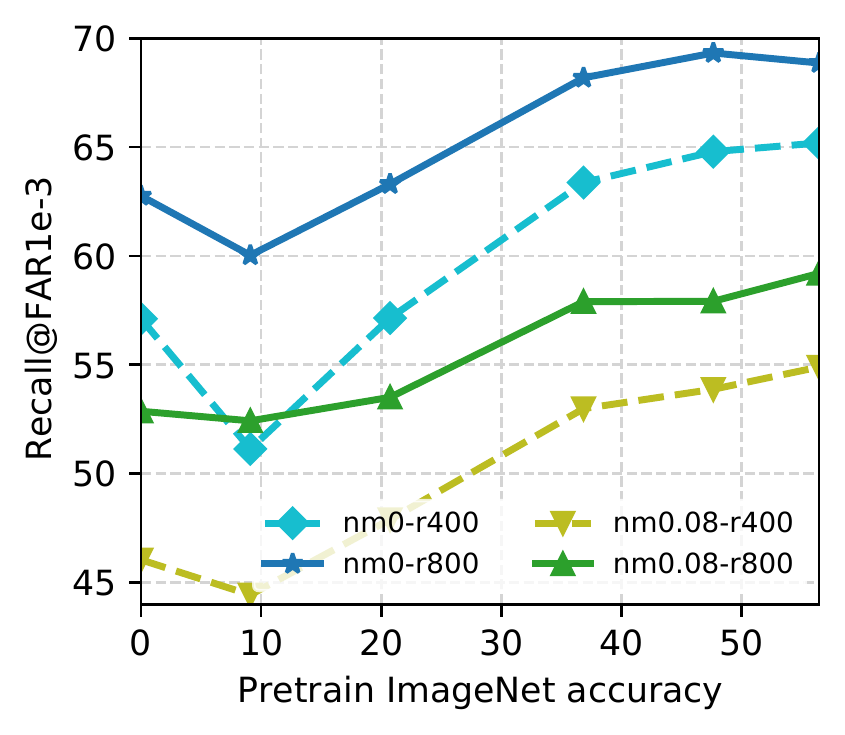}
\caption{}
\label{fig:digi-abl-pretr}
\end{subfigure}
\caption{ \Ourmetric on DigiFace validation set when training a MobileNet (a)  with \ouralg w/ pretraining, and baselines of \ouralg w/o pretraining and DP-\fedavg; (b) with different pretrained models. 
}
\label{fig:digi-freeze}
\ifthenelse{\boolean{cvpr}}{\vspace{-0.3cm}}{}
\end{figure}

\textbf{Parameter freezing and public pretraining.}
For similar parameter size, MobileNetV2 outperforms ResNet-50 with frozen parameters, and \cref{fig:digi-mobilenet-noise} shows the privacy-utility trade-off. \Ourmetric of \ouralg-r800 on MobileNetV2 only drops from $68.86\%$ to $67.56\%$ when $0.02$ noise is added, while ResNet-50 drops from $79.09\%$ to $72.6\%$. However, ResNet-50 still outperforms MobileNetV2 by a large margin in the high utility regime. \ouralg consistently outperforms DP-\fedavg when noises are added. In \cref{fig:digi-abl-pretr}, though the private fine-tuning utility is not linearly increasing with pretraining accuracy, there seems to be a general positive correlation: better pretrained models can lead to better private models except for one outlier where a inferior pretrained model causes difficulty in training. More discussion on parameter freezing and public pretraining is provided in \cref{sec:app_param_freeze}.

%% file: exp_param_freeze.tex
\begin{figure*}[htb]
\centering
\begin{subfigure}[b]{0.21\textwidth}
\centering
\includegraphics[width=\textwidth]{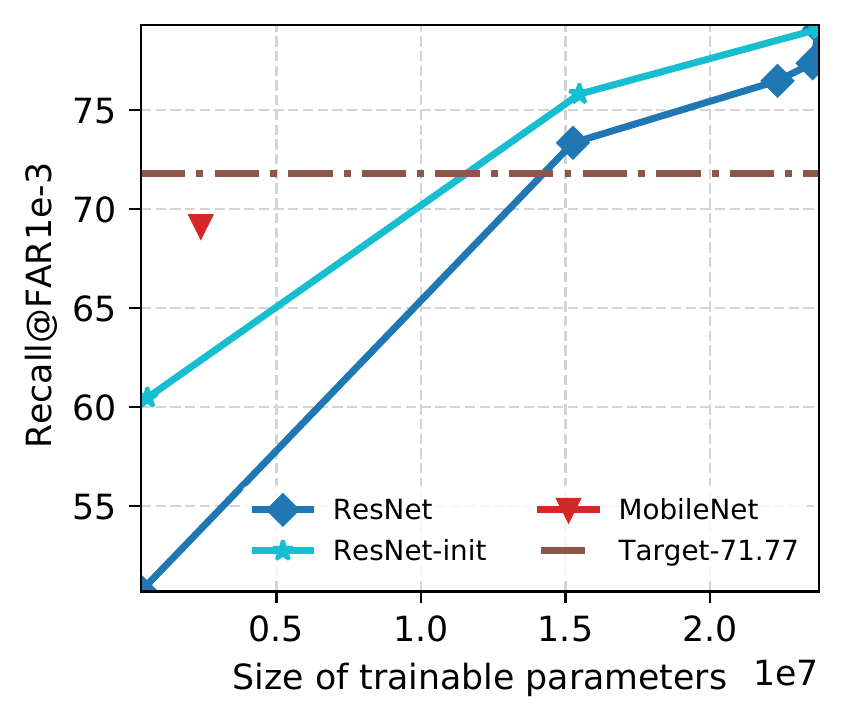}
\caption{}
\label{fig:digi-freeze-util}
\end{subfigure}
\begin{subfigure}[b]{0.23\textwidth}
\centering
\includegraphics[width=\textwidth]{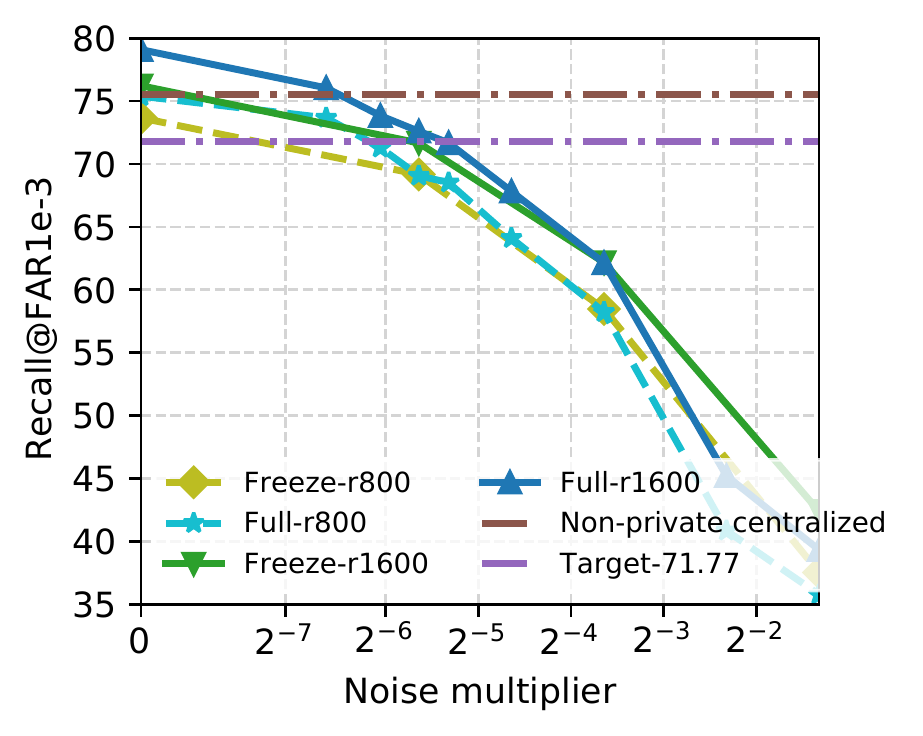}
\caption{}
\label{fig:digi-freeze-noise}
\end{subfigure}
\begin{subfigure}[b]{0.25\textwidth}
\centering
\includegraphics[width=\textwidth]{figs2/digi-mobilenet-noise.pdf}
\caption{}
\label{fig:digi-mobilenet-noise}
\end{subfigure}
\begin{subfigure}[b]{0.21\textwidth}
\centering
\includegraphics[width=\textwidth]{figs2/digi-abl-pretr.pdf}
\caption{}
\label{fig:digi-abl-pretr}
\end{subfigure}
\caption{ \Ourmetric on the DigiFace validation set when training (a) a ResNet with partial frozen parameters for $800$ rounds without noise;
(b) a partially frozen ResNet with \ouralg; (c) a MobileNet with \ouralg w/ pretraining, and baselines of DP-\fedavg and \ouralg w/o pretraining; (d) a MobileNet with different pretrained models. 
}
\label{fig:digi-freeze}
\end{figure*}

\input{exp_param_freeze_text}

%% file: exp_param_freeze_text.tex
\textbf{Parameter freezing.}
In \cref{fig:digi-noise-util,fig:digi-ftrl-noise-util}, we notice that the utility measured by \ourmetric can decrease faster when increasing the noise multiplier than observed for models in previous work \citep{kairouz21practical}. After using \ouralg to reduce the size of parameters to be noised, the ResNet-50 backbone still has ${\sim}24M$ parameters, which is $6 \times$ the language model in \citep{kairouz21practical} that has ${\sim}4M$ parameters. We explore freezing parameters and the alternative model architecture MobileNetV2 of ${\sim}2.4M$ parameters. We train parameters of all normalization layers, and gradually freeze the convolutional kernels from lower level to higher level (w or w/o the input convolutional layers) to generate \cref{fig:digi-freeze-util}. For image-to-embedding models, \ourmetric linearly increases with the size of parameters, which is different from the observation that models are redundant for image classification \citep{frankle2021training,sidahmed2021efficient}. Additionally training the input convolutional layers \citep{cattan2022fine} is more efficient than only training the higher levels of the network. In \cref{fig:digi-freeze-noise}, we freeze the convolutional kernels of intermediate two groups of residual blocks (out of the total four groups) in ResNet-50, which leads to a backbone network of $15.48M$ parameters. The partially frozen model is inferior to the full model for small-medium noise, and only effective in the low-utility regime of large noise $0.4$.

For similar parameter size, MobileNetV2 outperforms ResNet-50 with frozen parameters, and \cref{fig:digi-mobilenet-noise} shows the privacy-utility trade-off. \Ourmetric of \ouralg-r800 on MobileNetV2 only drops from $68.86\%$ to $67.56\%$ when $0.02$ noise is added, while ResNet-50 drops from $79.09\%$ to $72.6\%$. However, ResNet-50 still outperforms MobileNetV2 by a large margin in the high utility regime. DP-\fedavg is worse than \ouralg when noises are added. 

\textbf{Public pretraining.}
Even though the input image size of DigiFace is $112\times112$, different from the ImageNet pretraning image size of $224\times224$, the pretrained scale-invariant backbone can consistently improve the performance by $>5\%$ under the same noise level, as shown in \cref{fig:digi-mobilenet-noise}. Comparing curves of round 400 and round 800, the gain of public pretraining is larger when trained with a smaller number of rounds. We also pretrain a few different MobileNetV2 models on ImageNet by varying the total training epochs, and summarize the results in \cref{fig:digi-abl-pretr}. Though the private fine-tuning utility is not linearly increasing with pretraining accuracy, there seems to be a general positive correlation: better pretrained models can lead to better private models except for one outlier where a inferior pretrained model causes difficulty in training. Without private training, the \ourmetric of these pretrained models on DigiFace (with ImageNet validation accuracy) are smaller than $0.6\%$. Due to the domain difference, the utility of the pretrained model on DigiFace can be low, and it may not be consistent with the accuracy on ImageNet. For example, a pretrained MobileNetV2 can achieve $56.43\%$ accuracy on ImageNet while only $0.27\%$ \ourmetric on DigiFace, but it can boost the \ourmetric of training with \ouralg and $0$ noise, from $57.12\%$ for round $400$ and $62.76\%$ for round $800$ to $65.19\%$ and $68.86\%$, respectively. Finally, pretraining may not always help. For example, when pretraining from the (preprocessed) Google Landmark (GLD) dataset, the final \ourmetric can be worse than without pretraining. 

%% file: exp_ablation_curves.tex
\begin{figure*}[htb]
\centering
\begin{subfigure}[b]{0.32\textwidth}
\centering
\includegraphics[width=\textwidth]{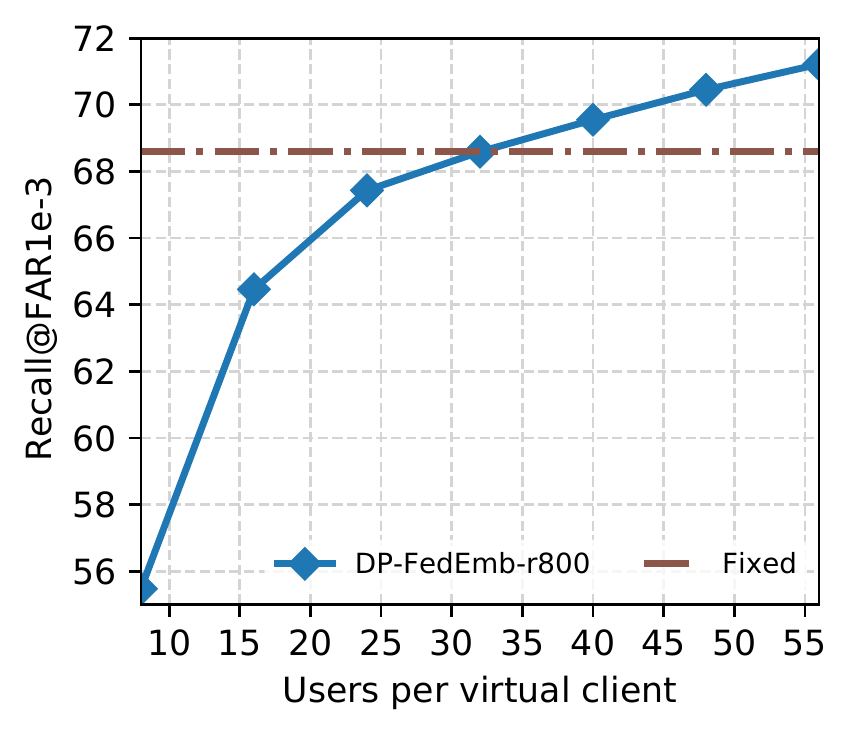}
\caption{}
\label{fig:digi-abl-dc}
\end{subfigure}
\begin{subfigure}[b]{0.32\textwidth}
\centering
\includegraphics[width=\textwidth]{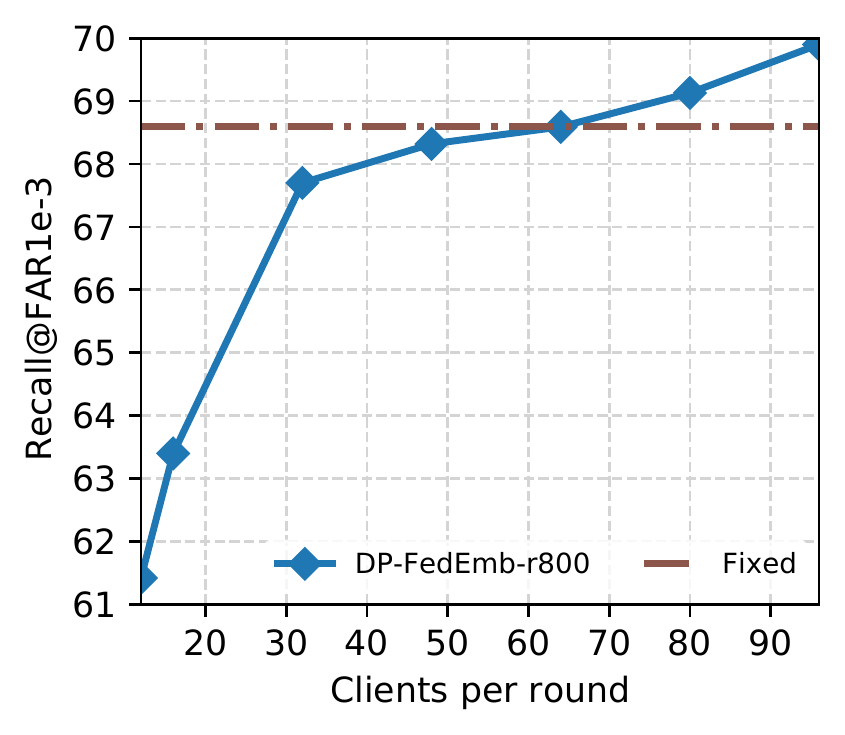}
\caption{}
\label{fig:digi-abl-cpr}
\end{subfigure}
\begin{subfigure}[b]{0.32\textwidth}
\centering
\includegraphics[width=\textwidth]{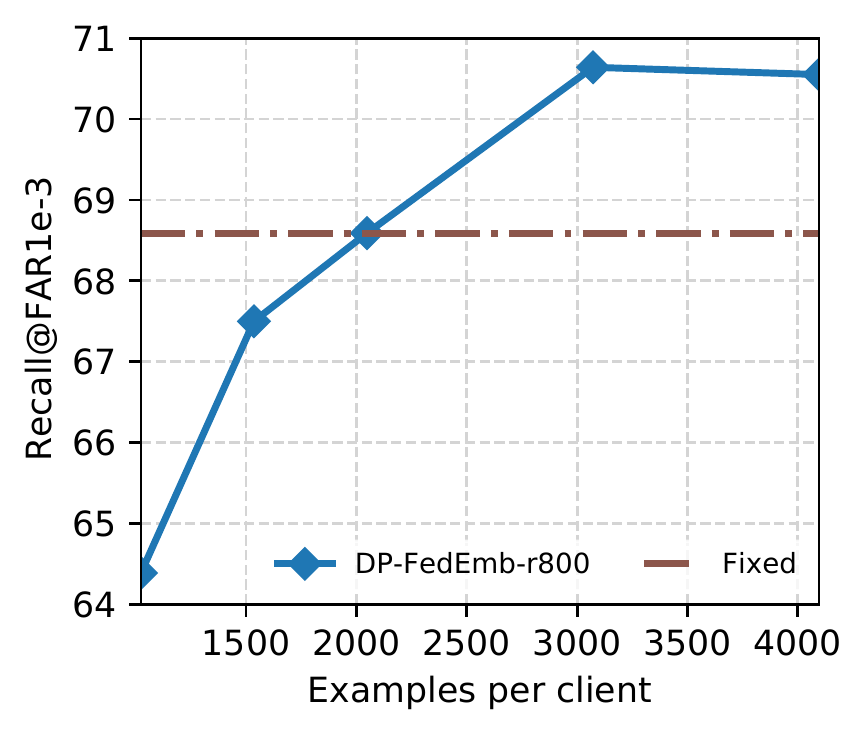}
\caption{}
\label{fig:digi-abl-epc}
\end{subfigure}
\begin{subfigure}[b]{0.32\textwidth}
\centering
\includegraphics[width=\textwidth]{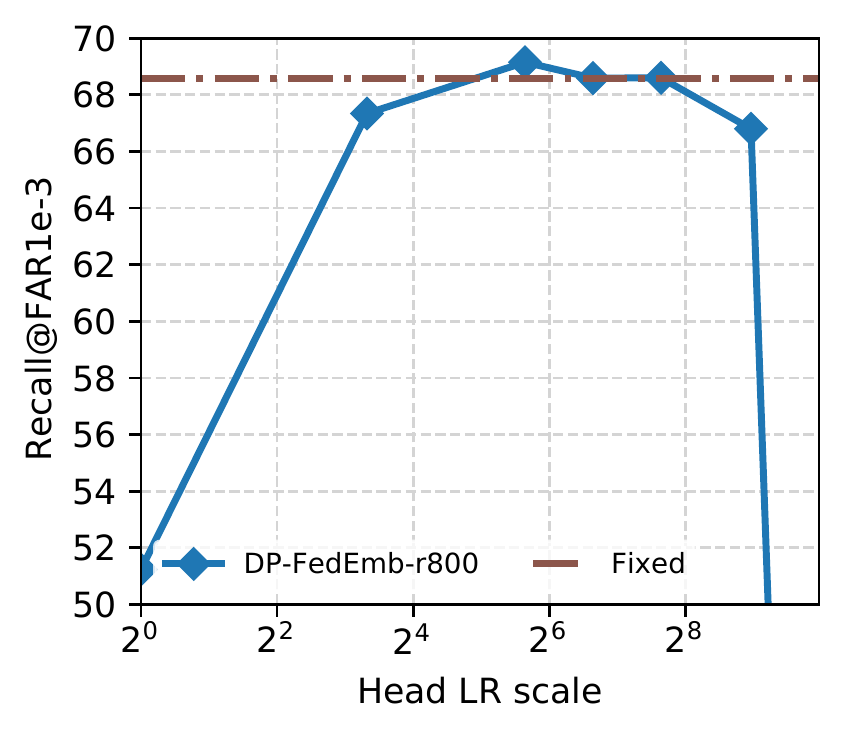}
\caption{}
\label{fig:digi-abl-hls}
\end{subfigure}
\begin{subfigure}[b]{0.32\textwidth}
\centering
\includegraphics[width=\textwidth]{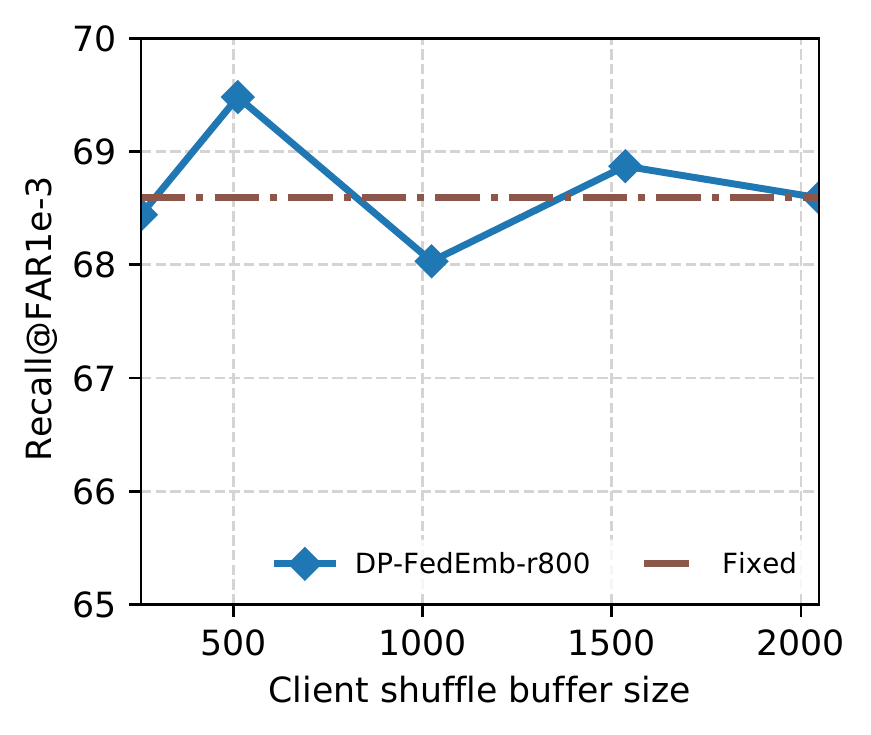}
\caption{}
\label{fig:digi-abl-csbs}
\end{subfigure}
\begin{subfigure}[b]{0.32\textwidth}
\centering
\includegraphics[width=\textwidth]{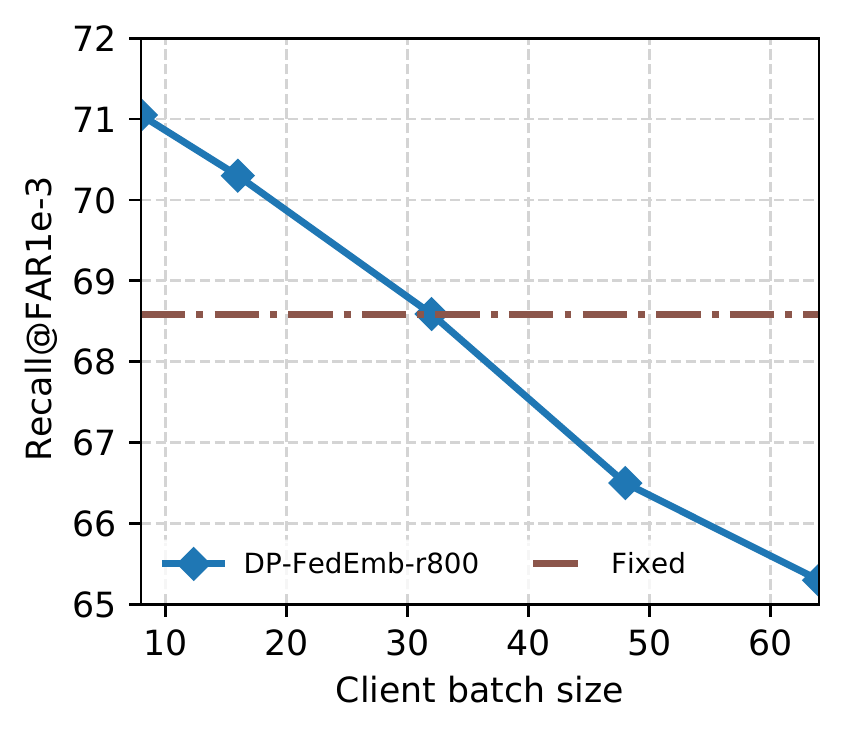}
\caption{}
\label{fig:digi-abl-cbs}
\end{subfigure}
\caption{ Ablation study for the fixed hyperparameters in (DP-)FedEmb; MobileNetV2 is trained for $800$ rounds on DigiFace10K with adaptive clipping \citep{andrew2019differentially} and zero noise; all the other hyperparameters are fixed when (a) users per dynamic client (b) clients per round (c) examples per client (d) head LR scale $\beta_2/\beta_1$ (e) client shuffle buffer size (f) client batch size is varied.}
\label{fig:ablation2}
\end{figure*}

%% file: exp_table_add_small.tex
{\small
\begin{table}[t]
    \centering
    \begin{tabular}{|c|c|c|c|}
    \hline
    \multirow{2}{*}{Dataset} & \multirow{2}{*}{Algorithm} & \multicolumn{2}{c|}{Recall@FAR=$1\mathrm{e}{-3}$ / $0.1$}  \\ 
    \cline{3-4} 
    &   & Approx  & AllPair  \\ 
    \hline
    \multirow{2}{*}{DigiFace} & DP-FedEmb  & - & $19.76\pm0.43$  \\
     & DP-FedAvg &    - &  $13.38\pm0.27$ \\
    \hline
     \multirow{2}{*}{EMNIST} & DP-FedEmb & $10.64\pm0.5$ & $10.47\pm0.5$\\
     & DP-FedAvg & $9.78\pm0.44$ & $9.67\pm0.41$ \\
    \hline
    \multirow{2}{*}{GLD} & DP-FedEmb  & $26.18\pm0.44$ & $27.07\pm0.04$ \\
     & DP-FedAvg &   $24.48\pm1.0$ &  $26.27\pm0.17$ \\
    \hline
    \multirow{2}{*}{iNat} & DP-FedEmb  &  $40.49\pm1.08$  & $40.93\pm0.94$  \\
     & DP-FedAvg &    $29.57\pm0.78$  & $29.6\pm0.65$ \\
    \hline
    \end{tabular}
    \caption{The utility under same privacy budget for DigiFace \citep{bae2023digiface1m}, EMNIST \citep{emnist}, GLD \citep{gldv2} and iNat \citep{inat} datasets.} 
    \label{tab:additional2}
\end{table}
}

%% file: exp_table_add.tex
\begin{table*}[htb]
    \centering
    \begin{tabular}{|c|c|c|c|c|c|c|c|}
    \hline
    \multirow{2}{*}{Dataset} & \multirow{2}{*}{Algorithm} & \multicolumn{4}{c|}{Hyperparameters} & \multicolumn{2}{c|}{Recall@FAR=$1\mathrm{e}{-3}$ / $0.1$}  \\ 
    \cline{3-8} 
    & & Noise & \text{SerLR} & \text{CliLR}  & \text{Clip} & Approx  & AllPair  \\ 
    \hline
    \multirow{2}{*}{DigiFace} & DP-FedEmb &  \multirow{2}{*}{$1.39$} & $5\mathrm{e}{-3}$ &\multirow{2}{*}{$2\mathrm{e}{-3}$} & $0.5$ & - & $19.76\pm0.43$  \\
     & DP-FedAvg &  & $2\mathrm{e}{-3}$ & & $1.5$ & - &  $13.38\pm0.27$ \\
    \hline
    \multirow{2}{*}{EMNIST} & DP-FedEmb & \multirow{2}{*}{$0.62$} & \multirow{2}{*}{$0.02$} & \multirow{2}{*}{$5\mathrm{e}{-3}$} & \multirow{2}{*}{$1$} & $10.64\pm0.5$ & $10.47\pm0.5$\\
     & DP-FedAvg & & & & & $9.78\pm0.44$ & $9.67\pm0.41$ \\
    \hline
    \multirow{2}{*}{GLD} & DP-FedEmb &  \multirow{2}{*}{$0.02$} & 1 &\multirow{2}{*}{$5\mathrm{e}{-4}$} & $0.3$ & $26.18\pm0.44$ & $27.07\pm0.04$ \\
     & DP-FedAvg &  & $0.5$ & & $0.7$ & $24.48\pm1.0$ &  $26.27\pm0.17$ \\
    \hline
    \multirow{2}{*}{iNat} & DP-FedEmb & \multirow{2}{*}{$0.5$} & $0.02$ & $5\mathrm{e}{-4}$  & $0.2$ &  $40.49\pm1.08$  & $40.93\pm0.94$  \\
     & DP-FedAvg &  & $0.01$ & $1\mathrm{e}{-3}$ & $1$  & $29.57\pm0.78$  & $29.6\pm0.65$ \\
    \hline
    \end{tabular}
    \caption{The utility under same privacy budget for DigiFace \citep{bae2023digiface1m}, EMNIST \citep{emnist}, GLD \citep{gldv2}, and iNat \citep{inat} datasets.} 
    \label{tab:additional2}
\end{table*}

%% file: sec_conclusion.tex
\section{Conclusion}
This paper presented \ouralg for training embedding models with user-level differential privacy.
We show how practical utility with strong privacy guarantees can be achieved in the data center, thanks to key algorithm design choices around the construction of virtual clients and in the selection of what information is shared among users.
Our experiments validate this improves the privacy utility trade-off upon vanilla DP-FedAvg for supervised representation learning.
Though strong formal DP bounds at practical levels of utility could only be achieved when millions of users participate in training, \ouralg is designed to be exceptionally scalable when model size and class space increases with number of users. \ouralg can also be applied to decentralized FL when each real client contains multiple classes, possibly reducing the necessity of virtual clients. Finally, DP is a worst-case guarantee that can be improved by both algorithmic design and advanced accounting methods; the non-negligible noise we added for the small scale datasets in experiments are ready to be empirically audited for privacy.  

%% file: sec_ack.tex
\subsection*{Acknowledgement}

The authors would like to thank Zachary Garrett, Keith Rush, and the TFF team for simulation support; Viral Carpenter and Janel Thamkul for support through the internal review process; Jun Xie and Lior Shapira for early discussion; and Peter Kairouz for early feedback.